\documentclass{article}
\usepackage{microtype}
\usepackage{graphicx}
\usepackage{booktabs}
\usepackage{tabularx}
\usepackage{array}
\usepackage{xurl}
\usepackage{hyperref}
\usepackage[preprint]{icml2026}
\usepackage[T1]{fontenc}
\usepackage[utf8]{inputenc}
\usepackage{amsmath,amssymb,mathtools}
\usepackage{placeins}
\usepackage{algorithm}
\usepackage{algorithmic}
\icmltitlerunning{Expanding the Decision Space of Pretrained Models}
\hypersetup{pdftitle={Expanding the Decision Space of Pretrained Models through Structured Re-Decoding},
  pdfsubject={Structured prediction and constrained inference},
  pdfkeywords={structured re-decoding, frozen models, candidate recombination, QUBO}}

\begin{document}
\raggedbottom
\twocolumn[
\icmltitle{Expanding the Decision Space of Pretrained Models\\
through Structured Re-Decoding}
\hypersetup{pdftitle={Expanding the Decision Space of Pretrained Models through Structured Re-Decoding}}
\icmlsetsymbol{first}{\textdagger}
\icmlsetsymbol{corresponding}{*}
\begin{icmlauthorlist}
  \icmlauthor{Tinglan Wang}{1,first}
  \icmlauthor{Qun Su}{2,3,first}
  \icmlauthor{Shihang Wang}{4,first}
  \icmlauthor{Chi Wang}{5,first}
  \icmlauthor{Lin Wang}{6}
  \icmlauthor{Shaobo Chen}{1}
  \icmlauthor{Haixu Liu}{1}
  \icmlauthor{Hai Wei}{1}
  \icmlauthor{Yin Ma}{1}
  \icmlauthor{Yu Kang}{2}
  \icmlauthor{Qi Gao}{1,corresponding}
\icmlauthor{Xiaojun Yao}{4,corresponding}
  \icmlauthor{Tingjun Hou}{2,corresponding}
  \icmlauthor{Kai Wen}{1,corresponding}
  
\end{icmlauthorlist}

\icmlaffiliation{1}{Beijing QBoson Quantum Technology Co., Ltd., Beijing, China}
\icmlaffiliation{2}{College of Pharmaceutical Sciences, Zhejiang University, Hangzhou, China.}
\icmlaffiliation{3}{Shanghai Innovation Institute, Shanghai, China.}
\icmlaffiliation{4}{Faculty of Applied Sciences, Macao Polytechnic University, Macao, China.}
\icmlaffiliation{5}{Department of Chemical Engineering, Tsinghua University, Beijing, China}
\icmlaffiliation{6}{Center for AI and Computational Biology, Institute of Systems Medicine, Chinese Academy of Medical Sciences, Suzhou, China}
\icmlcorrespondingauthor{Qi Gao}{gaoq@boseq.com}
\icmlcorrespondingauthor{Xiaojun Yao}{xjyao@mpu.edu.mo}
\icmlcorrespondingauthor{Tingjun Hou}{tingjunhou@zju.edu.cn}
\icmlcorrespondingauthor{Kai Wen}{wenk@boseq.com}
\vskip 0.3in
]
\printAffiliationsAndNotice{}
\begin{abstract}

Native decoders limit how pretrained models combine local predictions under new structural constraints. We introduce Structured Re-Decoding (SRD), a training-free framework that constructs local candidates from model scores, samples, or external specifications, jointly optimizes their selection, and verifies reconstructed outputs. We show that candidate recombination expands feasible support and can improve attainable reward under a fixed KL divergence budget. Across molecular generation, protein design, RNA structure prediction, and antibody modeling, SRD enables improved structured inference, including verified molecules absent from the original sample pool. External chemical templates further extend SRD to noncanonical amino acids, peptide cyclization, and grammar-guided molecular construction. Preliminary experiments on a physical coherent Ising machine demonstrate the feasibility of hardware-based optimization, suggesting broader opportunities for quantum and quantum-inspired approaches to structured inference and combinatorial optimization across scientific domains.
\end{abstract}

\section{Introduction}
\label{sec:1}

\begin{figure*}[!t]
  \centering
  \includegraphics[width=\textwidth]{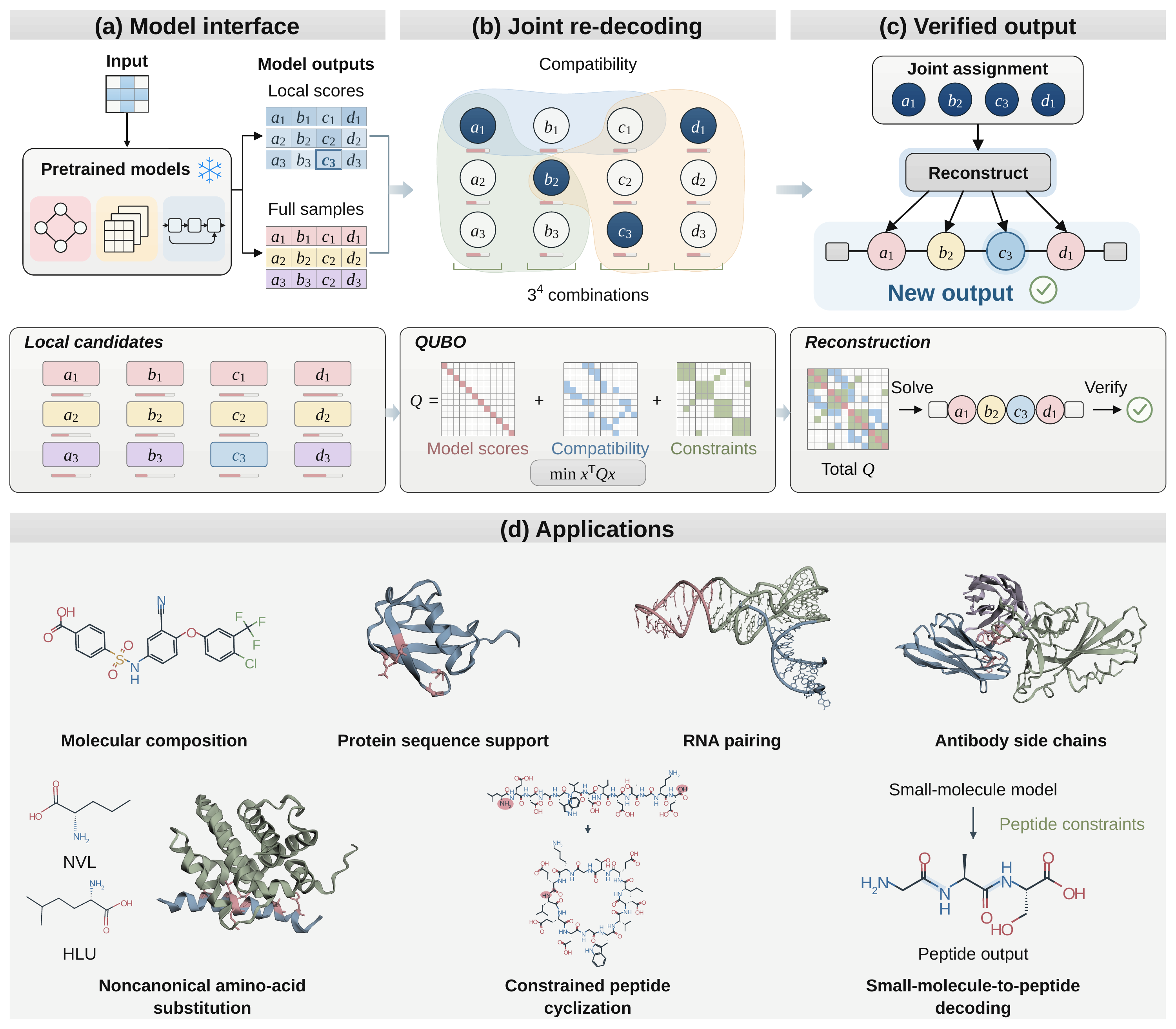}
  \caption{\textbf{Structured re-decoding of frozen pretrained models and its applications.}
  \textbf{(a) Model interface.} Local candidates come from prediction scores or aligned complete samples. Light-blue $c_3$ illustrates a score-derived alternative absent from the displayed samples.
  \textbf{(b) Joint re-decoding.} Model scores, pairwise compatibility, and constraint penalties form the QUBO matrix. Colored regions indicate interaction scopes; overlaps denote shared candidates. Green brackets group alternatives for each decision unit, with exactly one selected per group. The illustrated four groups of three states define $3^4$ assignments before constraints. Dark-blue nodes mark the selected assignment.
  \textbf{(c) Verified output.} The assignment is reconstructed into a complete object and checked against the task's acceptance policy. ``New output'' denotes an object outside the original complete-output pool, obtained by recombining observed states or using score-derived alternatives. The green check denotes acceptance.
  \textbf{(d) Applications.} The upper row presents molecular composition re-decoding (FLOWR), protein sequence-support analysis (ProteinMPNN), RNA pairing (KnotFold), and antibody side-chain coordination (DiffAb), from left to right. The lower row illustrates extensions to noncanonical amino-acid substitution, constrained peptide cyclization, and small-molecule-to-peptide decoding. NVL and HLU label the illustrated noncanonical residues. Structures illustrate the objects and operations studied.}
  \label{fig:workflow}
\end{figure*}

Screening a finite pool of complete model outputs limits delivery to objects the model has already produced. Local scores or observed states may expose additional combinations that satisfy a new task specification. Examples include molecules with a requested composition and RNA structures with mutually exclusive base pairs. These decisions must be coordinated because a locally preferred choice can conflict with choices elsewhere in the object.

Existing structured inference already coordinates learned scores under constraints~\citep{roth2004global}, including RNA pairing in IPknot and KnotFold~\citep{sato2011ipknot,gong2024knotfold}. Fusion Moves selects compatible labels from proposal solutions~\citep{lempitsky2010fusion}, and QSQS uses quadratic overlap costs to select detections~\citep{li2020qsqs}. These methods motivate our use of coupled selection. We study two ways to construct its inputs after a frozen model has run. Recombining aligned sample states extends the choices available from complete proposals. Using local prediction scores can also introduce alternatives absent from the proposals. The intervention acts on model outputs, whereas PRODIGY enforces constraints during frozen diffusion~\citep{sharma2024prodigy}.

We introduce Structured Re-Decoding (SRD), an inference-time framework that jointly selects local candidates under task constraints, reconstructs the selected structures, and verifies them against the requested specification (Figure~\ref{fig:workflow}). SRD expands the decision space in two ways. It recombines states available from model scores or samples, and it can incorporate external definitions that supply new candidate states or prescribe how an output is assembled. Both operate with the pretrained model's parameters fixed.

\begin{enumerate}
\item \textbf{Which decisions become available?} On six four-site ProteinMPNN domains~\citep{dauparas2022proteinmpnn}, exact enumeration quantifies the conditional probability mass, minimum KL divergence, and reward at a fixed KL budget available from recombining the same sampled states.
\item \textbf{Which additional outputs can be delivered?} On FLOWR~\citep{cremer2026flowr}, we compare verified molecular delivery from terminal-score alternatives with screening the same generated outputs. A shared objective lets us compare local and full-domain search.
\item \textbf{How do compatibility terms affect task metrics?} KnotFold compares pairing objectives on shared prediction scores. DiffAb~\citep{luo2022diffab} compares unary selection with a pairwise policy and uses a separate fixed-identity control to test rotamer coordination.
\item \textbf{How can a frozen model support a new structural task?} We combine model scores or samples with external chemical definitions to construct and verify noncanonical substitutions, cyclic peptides, and peptide structures from general molecular model outputs.
\end{enumerate}

\section{Structured Re-Decoding Method}
\label{sec:2}

\subsection{Candidate construction}
\label{sec:2.1}

Given an input context $c$, an adapter defines editable units, protected context, local candidate sets $\mathcal A_i$, and a deterministic reconstruction map $D_c$ for joint states $s\in\Omega=\prod_i\mathcal A_i$. Each candidate retains its unit identity, state, coordinates where applicable, and score provenance.

\textbf{Score-derived candidates} retain alternatives permitted by a frozen predictor. FLOWR exposes atom, charge, and bond choices from terminal predictions; KnotFold supplies scored pairing edges. These candidates can include states absent from every complete source output. \textbf{Sample-derived candidates} align decision units across complete outputs and retain observed states, as in the exact ProteinMPNN analysis. DiffAb combines score-derived amino-acid alternatives with constructed rotamers. An external task specification can supply candidate states or define constraints and reconstruction rules. For example, a residue library supplies noncanonical identities and rotamers, whereas a peptide grammar specifies the connectivity of the reconstructed object. The adapter records these external inputs separately from model-provided scores and sampled states. The support inclusion result in Section~\ref{sec:2.3} uses aligned sample-derived states.

\subsection{Joint objective and constraints}
\label{sec:2.2}

For multistate units, $x_{ia}=\mathbf1\{s_i=a\}$ indicates the selected candidate. Native binary decisions, such as RNA pairing edges, keep their binary representation. Unary costs $u_i$ encode model preferences, edits, or local geometry. Pairwise costs $v_{ij}$ coordinate interacting choices:

\begin{equation}
\label{eq:1}
\begin{aligned}
\widehat E(x)={}&e_0+\sum_{i,a}u_i(a)x_{ia}\\
&+\sum_{i<j}\sum_{a,b}v_{ij}(a,b)x_{ia}x_{jb}.
\end{aligned}
\end{equation}

The quadratic expression in Eq.~\eqref{eq:1} evaluates the chosen cost tables exactly. Its agreement with downstream metrics depends on the task. When a complete-state objective can be queried, anchored single and double changes give a quadratic approximation (Appendix~\ref{app:a}). Recombined joint model probabilities require their own evaluations.

Additive equalities $\sum_{i,a}d_{ria}x_{ia}=b_r$ and one-hot constraints can be expressed by squared penalties, giving a quadratic unconstrained binary optimization (QUBO) representation:

\begin{equation}
\label{eq:2}
\begin{aligned}
E_\rho(x)={}&\widehat E(x)
+\rho\sum_i\left(\sum_a x_{ia}-1\right)^2\\
&+\rho\sum_r\left(\sum_{i,a}d_{ria}x_{ia}-b_r\right)^2.
\end{aligned}
\end{equation}

The first penalty sum applies to multistate units. Inequalities may remain explicit or use bounded integer slack. If the domain admits a feasible assignment and every encoded violation contributes at least one, a penalty weight above the sum of absolute nonconstant soft coefficients makes every exact minimizer feasible. Appendix~\ref{app:a} gives the bound and its assumptions.

\subsection{Reconstruction, verification, and fallback}
\label{sec:verify}

The decoder may use the QUBO or an equivalent constrained formulation. Local search restricts updates to neighborhoods, full-domain search admits the entire declared candidate domain, and a certificate bounds the finite objective. Joint modeling can use any of these procedures. FLOWR compares two search scopes on one objective, KnotFold's stacking objective uses an equivalent sparse MILP, and the primary DiffAb comparison uses twelve coordinate sweeps.

The selected state is reconstructed as $y=D_c(s)$. The verifier distinguishes encoded feasibility, domain validity, and downstream utility. A task-specific policy ranks accepted candidates. When none is accepted, an anchor-based policy returns the original output with its actual validity and a fallback flag. FLOWR applies this policy to chemistry and composition. The primary DiffAb comparison measures sterics before its optional physical acceptance gate and retains construction-time fallbacks. Algorithm~\ref{alg:redecoding} and Appendix~\ref{app:b} specify the procedure and task protocols.

\begin{table*}[!t]
\centering
\caption{The four primary evaluations and their attribution. Full-domain search refers to the entire declared candidate domain. Objective certificates and downstream verification are tracked separately.}
\label{tab:attribution}
\small
\setlength{\tabcolsep}{4pt}
\begin{tabularx}{\textwidth}{@{}p{0.115\textwidth}p{0.17\textwidth}p{0.19\textwidth}X p{0.155\textwidth}@{}}
\toprule
Study & Candidate provenance & Joint decision & Baselines and attribution & Primary endpoint \\
\midrule
FLOWR & First output's terminal atom, charge, and bond scores & Bounded fixed-coordinate edits with composition targets & Best-of-four shares generation but changes the available decisions and uses screening. Local versus full-domain search shares support and objective. & Composition-and-quality delivery; overall quality pass \\
ProteinMPNN & Observed states from the same $K$ feasible native draws & Exact feasible product support on four sites & Whole-pool versus recombined support shares samples, constraints, reward, and conditional reference. & Retained mass, minimum KL, and reward frontier \\
KnotFold & Frozen pairing scores; eligibility depends on decoder & Pair exclusivity and an optional stacking reward & Canonical matching provides the zero-reward objective reference. Local and full-domain stacking share support and objective. & Reference macro F1 \\
DiffAb & Score-derived identities and constructed rotamers & Pairwise side-chain compatibility & Both arms share candidates. The pairwise arm adds a hard-pair fallback. A fixed-AA control holds identities fixed. & Specified steric endpoints before the optional gate \\
\bottomrule
\end{tabularx}
\end{table*}

\subsection{Support expansion and KL characterization}
\label{sec:2.3}

The sample-derived interface permits an exact mechanism analysis when model probabilities are computable. Relative to a specified model distribution $p_0$ and a hard-feasible set $\mathcal C$ with positive mass, define $p_C(s)=p_0(s)\mathbf1_{\mathcal C}(s)/p_0(\mathcal C)$. From a fixed native stream, take the first $K$ feasible samples and deduplicate them into $\mathcal S_{\rm pool}$. Duplicates consume the sampling budget and are not replaced. If $\mathcal A_i^{(K)}$ contains the states observed at position $i$, then $\mathcal S_{\rm rec}=(\prod_i\mathcal A_i^{(K)})\cap\mathcal C$ contains $\mathcal S_{\rm pool}$. Both supports share a generation budget, while their subsequent computation costs can differ. The following standard identities quantify the probability mass and utility made accessible by these particular supports.

For any nonempty support $\mathcal S\subseteq\mathcal C$ with positive reference mass and any distribution $\nu$ supported there,

\begin{equation}
\label{eq:3}
\begin{aligned}
D_{\rm KL}(\nu\Vert p_C)
={}&D_{\rm KL}\!\left(\nu\Vert p_C(\cdot\mid\mathcal S)\right)\\
&-\log p_C(\mathcal S).
\end{aligned}
\end{equation}

Substituting $p_C(s\mid\mathcal S)=p_C(s)/p_C(\mathcal S)$ into the log ratio proves the identity. By nonnegativity of KL, Eq.~\eqref{eq:3} implies that the conditional reference achieves the minimum attainable divergence $-\log p_C(\mathcal S)$. This standard identity assigns a divergence cost to the probability mass omitted by a restricted pool. Using the common reference retains that cost in comparisons across supports.

For reward $r$ and a KL budget $\varepsilon$ in nats, define

\begin{equation}
\label{eq:4}
F_{\mathcal S}(\varepsilon)=
\sup_{\substack{\operatorname{supp}(\nu)\subseteq\mathcal S\\
D_{\rm KL}(\nu\Vert p_C)\leq\varepsilon}}
\mathbb E_\nu[r].
\end{equation}

For the frontier in Eq.~\eqref{eq:4}, set inclusion gives $F_{\mathcal S_{\rm rec}}(\varepsilon)\geq F_{\mathcal S_{\rm pool}}(\varepsilon)$ whenever the pool is feasible. The added states, their probabilities, and their rewards determine whether the improvement is strict. Exponential tilting computes the finite-support frontier, as derived in Appendix~\ref{app:a}. Four-site enumeration measures this opportunity exactly. Reaching the frontier with a finite-time sampler requires a separate evaluation.

\section{Experimental Setup}
\label{sec:3}

The four primary evaluations ask complementary questions about delivery, accessible support, and structural coordination. Table~\ref{tab:attribution} identifies the candidate source and what each comparison changes. Appendix~\ref{app:b} gives the detailed comparisons and application protocols; Appendix~\ref{app:d} gives the exact ProteinMPNN protocol.

FLOWR uses 1,800 requests across 225 holo pockets and eight seeds, with four complete generations per request. Its single target is N+O=4; the dual target adds one halogen. Success requires composition and the declared RDKit/PoseBusters checks~\citep{buttenschoen2024posebusters}. ProteinMPNN enumerates four editable positions with 20 standard residues on each of six backbones, with four native streams per backbone. Charge and hydrophobic count define feasibility. Aromatic and polar count losses define rewards (Appendix~\ref{app:d}).

RNA uses 7,215 records in six UFold test files~\citep{fu2022ufold} and compares decoders on common KnotFold scores. DiffAb uses 3,600 cases from 50 complexes, with 72 outputs nested within each complex. Paired intervals cluster FLOWR by pocket, RNA by sequence hash, and DiffAb by complex. Planned denominators retain failures. The studies were developed in stages; known RNA training and development overlap and antibody proxy-relatedness are reported in Appendix~\ref{app:b}.

\section{Main Results}
\label{sec:4}

\begin{figure*}[!t]
  \centering
  \includegraphics[width=\textwidth]{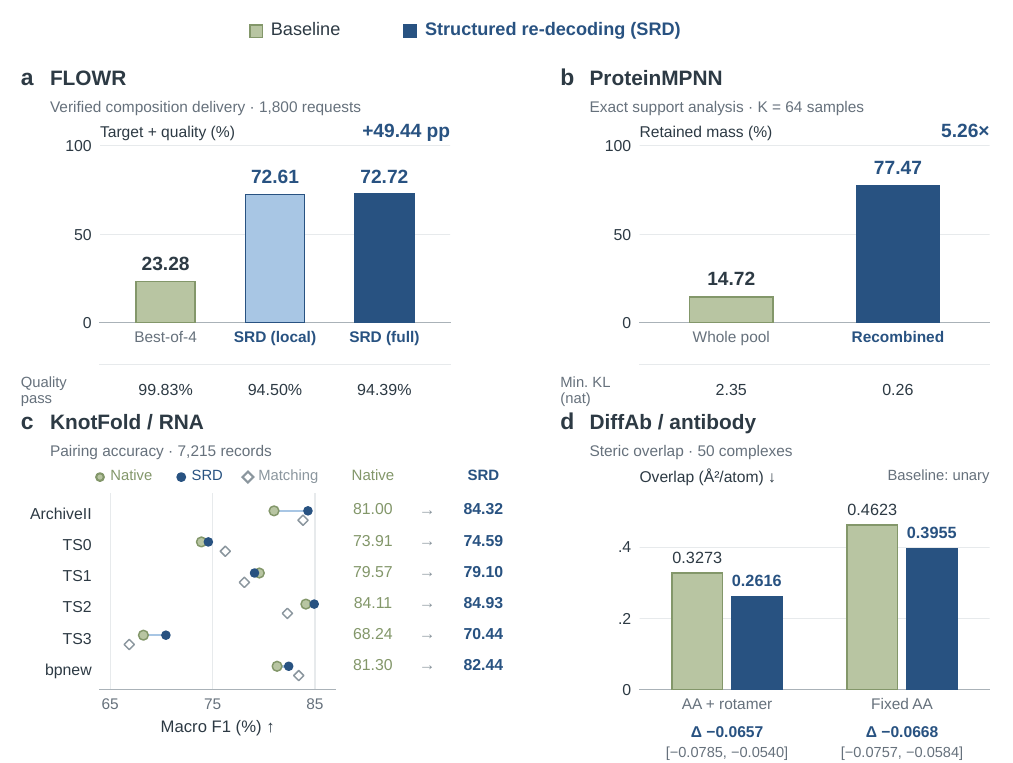}
  \caption{\textbf{Structured re-decoding (SRD) compared with task-specific baselines.}
  \textbf{(a)} FLOWR single-target success requires composition and quality checks, over 1,800 requests with matched source-generation budgets. Local and full-domain SRD optimize the same objective with different search scopes. The gain is full-domain minus best-of-four success; the lower row reports quality passage under each fallback policy.
  \textbf{(b)} Exact ProteinMPNN support analysis using the same $K=64$ feasible draws. Bars show mean conditional mass; the lower row gives mean per-stream minimum KL across six four-site domains, each with four streams. The ratio compares mean masses.
  \textbf{(c)} Native KnotFold flow and full-domain stacking SRD are connected within each dataset, with absolute macro F1 listed alongside. Gray diamonds show canonical matching. Table~\ref{tab:2} reports paired changes and 95\% sequence-cluster intervals.
  \textbf{(d)} Mean DiffAb overlap before and after pairwise SRD. The baseline is unary selection. Below each group, $\Delta$ is SRD minus unary with its paired 95\% complex-cluster interval (50 complexes, 3,600 cases). The AA-and-rotamer policy uses a hard-pair fallback; the fixed-AA control uses a sequence-preserving fallback.}
  \label{fig:evidence}
\end{figure*}

Figure~\ref{fig:evidence} summarizes the four primary evaluations. We first measure verified delivery from terminal scores, then quantify the separate sample-support mechanism, and finally test structural coordination in prediction and geometry. Section~\ref{sec:peptide_capabilities} extends the interface to peptide alphabets and topology.

\subsection{FLOWR: Verified constrained molecular delivery}
\label{sec:4.1}

Full-domain re-decoding increases FLOWR's composition-and-quality success from 419 to 1,309 of 1,800 requests for the single target, and from 119 to 615 for the dual target, relative to best-of-4 selection (Table~\ref{tab:1}). These gains are 49.444 percentage points (95\% pocket-cluster interval [45.167, 53.611]) and 27.556 points [24.056, 31.056]. Each request has four generated outputs across 225 pockets and eight seeds. Best-of-four screens the four complete outputs. Re-decoding instead uses the first output's terminal scores to construct editable states. The comparison fixes the source-generation budget and measures the effect of the full delivery policies.

\begin{table*}[t]
\centering
\caption{FLOWR delivery across all 1,800 scheduled requests. Success requires both the composition target and the specified quality checks; percentages use the full request denominator. The single target is N+O=4; the dual target also requires one halogen. Coordinates remain fixed with at most eight changed atom, charge, or bond decisions. Quality uses the declared RDKit and PoseBusters 0.6.5 \texttt{mol\_fast} checks.}
\label{tab:1}
\small
\setlength{\tabcolsep}{4pt}
\begin{tabularx}{\textwidth}{@{}lXrr@{}}
\toprule
Target & Method & Success, n (\%) & Quality pass, n (\%) \\
\midrule
Single & Original output & 143 (7.944) & 1,619 (89.944) \\
Single & Best-of-4 & 419 (23.278) & 1,797 (99.833) \\
Single & Local quadratic search & 1,307 (72.611) & 1,701 (94.500) \\
Single & Full-domain re-decoding & 1,309 (72.722) & 1,699 (94.389) \\
Dual & Original output & 36 (2.000) & 1,619 (89.944) \\
Dual & Best-of-4 & 119 (6.611) & 1,797 (99.833) \\
Dual & Local quadratic search & 615 (34.167) & 1,696 (94.222) \\
Dual & Full-domain re-decoding & 615 (34.167) & 1,694 (94.111) \\
\bottomrule
\end{tabularx}
\end{table*}

With a first-output fallback, full-domain re-decoding passes the quality checks on 94.389\% of single-target requests and 94.111\% of dual-target requests. Best-of-four screening passes on 99.833\% for both targets. These are outcomes of the tested delivery policies. Appendix~\ref{app:b} gives bounds for a policy that can retain either arm's output. Local quadratic search achieves nearly the same success as full-domain search. The differences are 0.111 points [\ensuremath{-}0.167, 0.389] for the single target and 0 points [\ensuremath{-}0.278, 0.278] for the dual target. This comparison measures the combined effect of terminal-score candidate construction, composition guidance, and verified selection.

The archive records 1,359 single-target and 1,260 dual-target deliveries outside their four-sample pools. By inclusion--exclusion, at least 868 single-target and 75 dual-target deliveries both succeed and fall outside the recorded pools. This bound describes selected outputs relative to the original samples. The next experiment isolates support expansion by recombining states observed in the same complete samples.

\subsection{ProteinMPNN: Quantifying candidate support expansion}
\label{sec:4.4}

Local recombination expands the available choices in the four-site protein domains. Each backbone has 160,000 enumerated states, of which 27,648 satisfy the hard constraints. For each of six backbones, four native streams supply the first $K$ feasible samples from 4,096 raw draws; repeated sequences are not replenished. Whole-pool selection and local recombination use those same samples and the same full conditional reference $p_C$. At $K=64$, recombination expands mean support from 57.67 to 8,041.79 states and retained reference mass from 14.7158\% to 77.4694\% (Table~\ref{tab:3}). Expanded support thus recovers model probability mass as well as distinct states.

\begin{table*}[t]
\centering
\caption{Support formed from the same $K$ feasible native samples. Values average within each backbone and then equally across six backbones; the 24 streams are nested replicates. Minimum KL is the mean of per-stream $-\log p_C(\mathcal S)$, not the negative logarithm of mean retained mass.}
\label{tab:3}
\small
\setlength{\tabcolsep}{4pt}
\begin{tabularx}{\textwidth}{@{}rXrrr@{}}
\toprule
$K$ & Support & Distinct states & $p_C$ mass (\%) & Min. KL (nat) \\
\midrule
4 & Whole pool & 3.96 & 2.0679 & 5.405006 \\
4 & Local recombination & 37.46 & 8.1172 & 3.612204 \\
16 & Whole pool & 15.58 & 6.1702 & 3.505393 \\
16 & Local recombination & 826.96 & 33.9579 & 1.184234 \\
64 & Whole pool & 57.67 & 14.7158 & 2.351939 \\
64 & Local recombination & 8,041.79 & 77.4694 & 0.258796 \\
\bottomrule
\end{tabularx}
\end{table*}

At $K=64$ and a KL budget of 0.5 nat, all 24 recombination supports admit a feasible distribution, whereas none of the whole pools does. At 4 nat, both supports are feasible for every stream; recombination strictly improves optimal dual-target reward in 23 streams, with an equal-backbone mean gain of 0.389288. Infeasible points are excluded from reward comparisons, rather than assigned zero utility. By support inclusion, every distribution feasible on the whole pool is also feasible on the recombined support. The observed strict gains quantify the value of the added states under the common reward and reference.

These finite-domain optima quantify attainable mass and reward under the declared count objectives, using model probabilities evaluated under each sequence's own prefixes. Appendix~\ref{app:d} separately reports full-domain target tilting and finite-budget probability-corrected sampling.

\subsection{KnotFold: Coordinating pairing predictions}
\label{sec:4.2}

With frozen KnotFold scores, full-domain stacking improves macro F1 over independent positive-score pairing on all six files (Table~\ref{tab:2}). The independent baseline permits multiple partners per nucleotide, so this comparison includes the benefit of resolving inconsistent local decisions. Relative to native flow, full-domain stacking improves five point estimates; ArchiveII, TS0, and bpnew have sequence-cluster intervals above zero. The gains over native flow extend the improvement from formal constraint satisfaction to reference pairing accuracy on some datasets.

\begin{table*}[t]
\centering
\caption{RNA macro F1 (\%) over all 7,215 records. The final column reports full-domain stacking minus native flow in percentage points with 95\% sequence-hash-cluster bootstrap intervals. File sizes are shown in parentheses; files are evaluated separately.}
\label{tab:2}
\small
\setlength{\tabcolsep}{4pt}
\begin{tabularx}{\textwidth}{@{}lrrrrrX@{}}
\toprule
File (n) & Independent & Native flow & Canonical & Local stack & Full stack & Full$-$native (pp) \\
\midrule
ArchiveII (393) & 72.525 & 81.004 & 83.835 & 84.298 & 84.317 & +3.313 [2.877, 3.739] \\
TS0 (1,305) & 68.790 & 73.908 & 76.247 & 74.765 & 74.585 & +0.677 [0.279, 1.073] \\
TS1 (60) & 75.660 & 79.568 & 78.107 & 79.222 & 79.098 & \ensuremath{-}0.470 [\ensuremath{-}1.570, 0.800] \\
TS2 (38) & 80.892 & 84.113 & 82.308 & 84.649 & 84.929 & +0.816 [\ensuremath{-}1.129, 3.119] \\
TS3 (18) & 66.365 & 68.241 & 66.855 & 68.808 & 70.438 & +2.196 [\ensuremath{-}1.379, 6.826] \\
bpnew (5,401) & 76.446 & 81.299 & 83.419 & 82.499 & 82.436 & +1.137 [0.978, 1.297] \\
\bottomrule
\end{tabularx}
\end{table*}

Canonical matching provides a zero-stacking-reward reference on the same admissible base types and separations. With optional pairing, adding negative-score edges cannot improve the linear matching objective (Appendix~\ref{app:b}). Relative to this reference, full-domain stacking improves ArchiveII by 0.482 points [0.141, 0.808] and reduces F1 on TS0 by 1.662 points [1.296, 2.028] and on bpnew by 0.983 points [0.833, 1.134]. Table~\ref{tab:s1} reports all six paired contrasts. Tie-breaking can change predictions without changing the matching objective.

The finite stacking problems achieved numerical optimal termination under the recorded solver settings. Local stacking has higher F1 point estimates on TS0, TS1, and bpnew despite using the same objective and candidates. The comparison shows that solving this objective to numerical optimality need not yield the highest reference F1. It does not distinguish changes due to strictly higher objective values from those due to tied solutions.

On the 591 TS0 records without a training-name match, full-domain stacking improves on native flow by 0.617 points [0.104, 1.143], while canonical matching remains stronger. On the 17 TS2 records outside development, local and full-domain stacking coincide and differ from native flow by 0.544 points [\ensuremath{-}2.189, 3.579]. Table~\ref{tab:rna_exposure} uses the documented name and development checks. Three ArchiveII records share sequence hashes with name-flagged TS0 records, and independence at the sequence-homology level remains unassessed.

\subsection{DiffAb: Compatibility-aware antibody re-decoding}
\label{sec:4.3}

On 50 DiffAb complexes with 3,600 nested cases, the policy with pairwise terms and a hard-pair fallback improves all four primary geometric endpoints relative to unary selection before the optional physical gate. The pairwise arm uses up to twelve coordinate sweeps. Normalized overlap falls from 0.327344 to 0.261613 \AA{}\textsuperscript{2} per atom, with paired difference \ensuremath{-}0.065732 [\ensuremath{-}0.078456, \ensuremath{-}0.054008]. Clashes per thousand evaluated CDR heavy atoms fall from 320.733 to 282.859, a difference of \ensuremath{-}37.874 [\ensuremath{-}44.411, \ensuremath{-}31.643]. Intervals cluster by complex. Per CDR, squared overlap within the CDR decreases by 4.88221~\AA{}\textsuperscript{2}, and overlap with the fixed context increases by 0.06676~\AA{}\textsuperscript{2}. Appendix~\ref{app:b} reports all endpoints. These checks quantify steric overlap and distances on the reconstructed structures.

The primary comparison changes residue identity in 762 cases. Of 1,010 substitutions, 725 reduce side-chain heavy-atom count; the mean count decreases by 0.659 atoms per CDR. An archived fixed-identity control on the same 50 complexes tests rotamer coordination directly. With the emitted DiffAb sequence fixed, pairwise selection reduces normalized overlap by 0.066829 \AA{}\textsuperscript{2}/atom [0.058381, 0.075667] across all 3,600 pairs. Both arms preserve the emitted identities, and rejected pairwise assignments return the unary output. The result measures the fixed-identity policy (Table~\ref{tab:cdr_fixed}).

FlowPacker provides a complementary fixed-identity cohort with a stronger packing anchor. Across 27 complexes and 648 cases, the protected policy accepts 35 recombinations and restores best-of-four in 613; mean clashes per thousand atoms decrease by 1.812 [$-2.896$, $-0.836$]. The primary DiffAb effect also persists in the 37 complexes below 90\% identity to the audited Fv proxy: normalized overlap changes by $-0.05867$ [$-0.07285$, $-0.04590$] (Table~\ref{tab:cdr_exposure}). These are proxy-relatedness strata, with checkpoint training membership unresolved.

\paragraph{Physical CIM execution.}
The antibody objectives also run on a physical coherent Ising machine (CIM). The best of ten retained CIM states improves the raw Ising objective over frozen fast simulated annealing (FSA-10) on 48/50 matrices of 177--521 spins and all seven enlarged matrices of 1,021--2,446 spins (Table~\ref{tab:cim_antibody}). Objectives are recomputed from archived integer couplings and returned spins. Common decoding and local polishing yield mostly tied molecular objectives, separating the raw optimization result from its downstream effect. The enlarged set comes from two source complexes. Appendix~\ref{app:antibody_cim} gives the experimental protocol, decoding procedure, and timing records.

\begin{table}[htbp]
\centering
\caption{Physical CIM execution of antibody objectives. Entries are CIM wins/ties/losses versus FSA-10. Each solver contributes ten retained states; physical trial budgets differ. Raw comparison maximizes $s^\top Js$; decoded comparison minimizes the molecular objective after strict restoration, repair, and common polishing.}
\label{tab:cim_antibody}
\small
\setlength{\tabcolsep}{4pt}
\begin{tabularx}{\columnwidth}{@{}Xrrr@{}}
\toprule
Matrices & Spins & Raw & Decoded \\
\midrule
Main (50) & 177--521 & 48/2/0 & 0/49/1 \\
Enlarged (7) & 1,021--2,446 & 7/0/0 & 1/6/0 \\
\bottomrule
\end{tabularx}
\end{table}

\subsection{Chemical alphabet and topology extensions}
\label{sec:peptide_capabilities}
SRD combines frozen model outputs with external chemical libraries and
topology rules to construct user-specified peptide structures. For
noncanonical substitution, ProteinMPNN provides residue-family
preferences, and an external library supplies noncanonical identities
and rotamers. For cyclization, PepGLAD supplies linear conformations,
and the task specifies head-to-tail closure. For peptide construction,
SRD maps atom roles defined by a peptide grammar to molecular atom
slots from FlowMol3. Table~\ref{tab:peptide_main} reports the outputs
that pass the chemical and geometric checks for each study.
Appendix~\ref{app:peptide_extensions} details the protocols.

\begin{table}[htbp]
\centering
\caption{Peptide capability outputs from QUBO re-decoding. Counts use the stated unit and each study's chemical and geometric checks. Fresh coordinates use ETKDGv3/MMFF94s.}
\label{tab:peptide_main}
\small
\setlength{\tabcolsep}{4pt}
\begin{tabularx}{\columnwidth}{@{}p{0.35\columnwidth}rX@{}}
\toprule
Extension & Qualified & Counting unit \\
\midrule
Noncanonical AA & 8/8 & Selected designs; one complex \\
Linear to cyclic & 2/4 & Source pools; one sequence--receptor pair \\
Linear tripeptide & 19/32 & Requests; fresh coordinates \\
Cyclic tetrapeptide & 28/32 & Requests; fresh coordinates \\
\bottomrule
\end{tabularx}
\end{table}

On the deposited PUMA--MCL1 backbone, ProteinMPNN supplies canonical side-chain-family preferences, while Rosetta NVL and HLU definitions supply noncanonical identities and rotamers~\citep{renfrew2012ncaa}. A 54-bit QUBO selects exactly two substitutions across six editable positions. All eight selected, chemically distinct designs pass independent graph, coordinate-derived stereochemistry, and contact checks. The external library expands the output alphabet; the canonical scores guide family selection within it.

For topology conversion, frozen PepGLAD's fixed-sequence checkpoint~\citep{kong2024pepglad} samples linear conformations of one supplied 13-residue sequence in the GABARAP environment. A 32-bit QUBO selects four mobile fragments, followed by GenKIC closure and constrained refinement. Across four pools of 32 source poses, three of 128 closure starts pass the declared chemical and geometric checks, yielding a qualified output in two pools. The resulting receptor-aligned RMSDs are 8.263--11.086~\AA{}; none reaches the 2~\AA{} near-native endpoint.

For general molecular outputs, FlowMol3~\citep{dunn2025flowmol3} supplies 32 unconditional endpoints. An external Gly/Ala/Ser/Val grammar enumerates size-compatible peptide templates, and a separate QUBO maps each template's atom roles onto the retained source slots. Fresh ETKDGv3/MMFF94s conformers yield verified linear tripeptides in 19/32 requests and cyclic tetrapeptides in 28/32. Relaxation from the original model coordinates yields 2/32 and 3/32, respectively.

\section{Discussion and Limitations}
\label{sec:5}

\paragraph{Candidate support.}
The ProteinMPNN enumeration isolates the effect of recombining observed states from the same samples. At a fixed KL budget, retaining more model probability mass lowers the minimum divergence and leaves more budget to favor the specified reward. The FLOWR result concerns a separate setting in which terminal scores define an edit domain and the decoder uses a composition objective and verification policy. Estimating the contribution of terminal scores in that setting requires a target-directed comparator with matched edit permissions and checks. For SemlaFlow, screening remains stronger on the same composition tasks (Appendix~\ref{app:c}).

\paragraph{Objective fidelity.}
KnotFold's numerical certificate applies to the stacking objective. The local and canonical decoders can still give higher reference F1. DiffAb's main policy reduces internal CDR overlap and slightly increases overlap with the context. The fixed-AA comparison shows a geometric benefit when the emitted sequence is held fixed. RNA training and development overlap and uncertainty about DiffAb checkpoint membership limit conclusions about independent generalization. Affinity, folding, stability, and biological function require their own measurements.

\paragraph{Search and cost.}
FLOWR's local search nearly matches full-domain delivery on the same objective. Matching generation budgets leaves candidate construction, optimization, and verification costs unmatched. The exact ProteinMPNN frontier describes what the tested support makes possible. Appendix~\ref{app:d} separately evaluates finite-budget sampling.

\paragraph{Peptide extensions.}
Section~\ref{sec:peptide_capabilities} shows how external chemical definitions compose with frozen model information to specify new residue alphabets and peptide topology. The delivered structures establish chemical construction and geometric verification under the declared checks. Appendix~\ref{app:peptide_extensions} gives the complete protocols and an additional fixed-chemistry thioether reconstruction study. Near-native pose recovery remains unresolved in both the cyclization and thioether settings.

\paragraph{Outlook for quantum optimization.}
The QUBO formulation of structured re-decoding provides a natural interface to emerging quantum and quantum-inspired optimization paradigms, including quantum annealing, gate-based quantum algorithms such as the Quantum Approximate Optimization Algorithm (QAOA), and coherent Ising machines (CIMs). Our preliminary experiments on a physical CIM demonstrate the feasibility of executing biologically motivated optimization problems on specialized Ising hardware. This unified formulation opens promising opportunities for integrating quantum and quantum-inspired optimization into structured inference and combinatorial optimization across diverse scientific domains. Future advances in quantum hardware and optimization algorithms may further expand the scale and complexity of coupled decisions that can be efficiently explored.

\section{Conclusion}
\label{sec:6}

Structured re-decoding combines local model alternatives into complete outputs under task constraints. FLOWR improves verified composition delivery under a fixed generation budget, and local search nearly matches full-domain search on the same objective. Exact ProteinMPNN enumeration shows how recombination recovers conditional probability mass and improves attainable reward in four-site domains. The RNA and antibody results show how pairing scores and steric objectives shape downstream gains. Together, these studies show how SRD expands the structures available from pretrained models through candidate recombination and uses external chemical definitions to address structural specifications beyond their native decoding interfaces.
\label{main:end}
\FloatBarrier

\bibliography{references}
\bibliographystyle{icml2026}

\clearpage
\appendix
\onecolumn
\setcounter{table}{0}
\renewcommand{\thetable}{S\arabic{table}}
\renewcommand{\theHtable}{supp.\arabic{table}}

\section{Candidate interfaces and constraint compilation}
\label{app:a}

\subsection{State identity, editing permissions, and score semantics}

An adapter associates each candidate with a stable decision-unit identity, its state value, the context required to reconstruct an output, and a score whose meaning is recorded. Recombination is defined only after that alignment. Residues, atoms, and layout elements retain their identities across the candidate arrays. Fixed coordinates, protected elements, and editing limits are part of the instance before optimization. The original output remains available as an anchor when the application uses selective correction.

The interface supports both categorical states and direct binary decisions. Antibody states represent amino-acid and side-chain choices with associated geometry. Molecular states represent permitted atom, charge, or bond choices under the fixed-coordinate editing contract. RNA variables directly indicate candidate base pairs, so there is no redundant one-hot group for each edge. Layout states associate a particular element with a candidate box. These adapters share one mathematical representation and a common output-checking interface. The antibody experiment uses its separately archived implementation.

Model logits, retrieval scores, geometric penalties, and learned pair potentials define optimization scores. Their probabilistic interpretation depends on the context in which they were computed. For the probability-aware extension, the interface recomputes the complete state probability under its own prefix, fixed context, temperature, alphabet, and decoding order; cached scores from other prefixes or trajectories remain optimization inputs.

\subsection{Penalties and verification}

Let the unpenalized upper-triangular binary polynomial have coefficient absolute sum $S$, excluding its constant. Each binary monomial changes by at most one between assignments, so the objective difference between any two assignments is at most $S$. Suppose the encoded feasible set is nonempty and an integer-valued violation polynomial $V$ is zero on feasible assignments and at least one otherwise. Then any coefficient $\rho>S$ makes every infeasible assignment worse than every feasible assignment for $E+\rho V$. The implementation uses the conservative choice $\rho=2S+1$ for this interface. Integer units, finite coefficients, and exact global optimization are essential conditions of this sufficient bound.

The bound applies to the encoded objective under exact optimization. Approximate solving, coefficient quantization, and conversion can alter feasibility, so returned states undergo checks for one-hot consistency, counts, editing permissions, and domain validity, including requirements evaluated outside the objective. Molecular evaluation uses the declared RDKit and PoseBusters checks; antibody evaluation uses the declared heavy-atom geometric measures. These endpoints assess physical validity; binding, stability, and clinical utility require their respective evaluations. This use of physical-validity checks follows~\citet{buttenschoen2024posebusters}.

A protected-output policy accepts a candidate only if all specified protection metrics satisfy their thresholds or do not deteriorate relative to the anchor, according to the frozen application protocol. If no acceptable replacement exists, the policy returns the anchor with its actual validity flag. An invalid anchor retains its failure status. Selective acceptance governs optimization delivery; probability-corrected sampling instead retains the chain output, including rejected transitions.

\subsection{Couplings and solver interpretation}

RNA stacking rewards adjacent stem pairs, while degree constraints prevent a nucleotide from participating in multiple selected pairs. The joint objective is solved through an equivalent sparse MILP. Its optimality certificate applies to the finite candidate objective; reference pairing accuracy is evaluated separately. Matched local search, matching, and native flow results are retained to separate domain information from solver effects.

When a full-state score $F$ is available, a quadratic surrogate can be constructed around an anchor $s^0$. A unary coefficient is $F(s^{i:a})-F(s^0)$, and a pair coefficient is

\[
F(s^{i:a,j:b})-F(s^{i:a})-F(s^{j:b})+F(s^0).
\]

The surrogate interpolates the queried anchor, single changes, and double changes; higher-order interactions are approximated by this quadratic representation. The protein center construction uses six chemical-class representatives at each of four sites, requiring $1+4\times5+6\times25=171$ true sequence scores. The surrogate fits the full-state score $F(s)=-\log p_0(s)$. The six classes are DE, KR, FWY, AVILM, STNQ, and CGHP. At each site, the representative is the most frequent calibration amino acid in its class, except that the first calibration sequence supplies the representative of its own class and the anchor state. Squared property losses are then added analytically. The subsequent sampler uses all 20 standard amino acids independently of the center-construction alphabet.

\subsection{Re-decoding procedure}

\begin{algorithm}[t]
\caption{Constrained re-decoding with task verification and anchor fallback}
\label{alg:redecoding}
\begin{algorithmic}[1]
\REQUIRE Frozen-model outputs and scores $O$, context $c$, anchor $y^0$, task specification $T$, solve budget $B$
\STATE Extract aligned candidates $\{\mathcal A_i\}$, provenance, and reconstruction map $D_c$ from $(O,T)$, including any library specified by $T$.
\STATE Build unary/pairwise costs and encoded constraints, including protected context and edit permissions.
\STATE Compile Eq.~\eqref{eq:2} or an equivalent constrained formulation; record solver scope.
\STATE Solve within $B$; collect assignments $X$ and solver statuses. Set accepted outputs $V\leftarrow\varnothing$.
\FOR{each $x\in X$}
\STATE Decode $x$ to $s$, reconstruct $y=D_c(s)$, and check feasibility, task validity, and protected metrics.
\IF{reconstruction succeeds and the declared policy accepts $y$}
\STATE Add $y$ and its scores/checks to $V$.
\ENDIF
\ENDFOR
\IF{$V\ne\varnothing$}
\STATE \textbf{return} the highest-ranked output in $V$ with verification and solve status.
\ELSE
\STATE \textbf{return} $y^0$ with its measured validity and fallback status.
\ENDIF
\end{algorithmic}
\end{algorithm}

\subsection{Finite-support reward and KL frontier}

Let $p_C$ be the common constraint-conditioned reference, and let $\mathcal S$ be a nonempty finite support with positive reference mass. For $t\geq0$, define $Z_{\mathcal S}(t)=\sum_{s\in\mathcal S}p_C(s)e^{t r(s)}$ and $q_t(s)=p_C(s)\mathbf1_{\mathcal S}(s)e^{t r(s)}/Z_{\mathcal S}(t)$. For any distribution $\nu$ on this support with finite $D_{\rm KL}(\nu\Vert p_C)$,

\[
D_{\rm KL}(\nu\Vert q_t)=D_{\rm KL}(\nu\Vert p_C)-t\mathbb E_\nu[r]+\log Z_{\mathcal S}(t).
\]

At an interior budget $\varepsilon=D_{\rm KL}(q_t\Vert p_C)$ with $t>0$, nonnegativity of the left side shows that any feasible $\nu$ has $\mathbb E_\nu[r]\leq\mathbb E_{q_t}[r]$. At $t=0$, $q_0=p_C(\cdot\mid\mathcal S)$ and the minimum feasible budget is $-\log p_C(\mathcal S)$. The reward derivative is $\operatorname{Var}_{q_t}(r)$ and the KL derivative is $t\operatorname{Var}_{q_t}(r)$, allowing a scalar search for interior budgets. At the maximum-reward plateau, the maximizers are taken over $\{s\in\mathcal S:p_C(s)>0\}$. The reference conditioned on those maximizers attains the plateau at the smallest feasible KL budget. Budgets below the minimum are infeasible.

The measured rewards are $-\ell_A$ for the single property and $-(\ell_A+\ell_P)$ for the two properties. Expected reward and probability of attaining both targets are different outcomes. Both supports are compared against the same $p_C$; conditioning the reference separately on each support would conceal the probability mass excluded by that support. Support inclusion gives weak frontier dominance by enlarging the feasible class of distributions. Strict gains in the experiment are empirical properties of its added states and model probabilities. This analysis applies standard KL identities and finite-domain variational optimization to quantify the effect of candidate support.

\section{Main application protocols and full geometric endpoints}
\label{app:b}

\begin{table}[htbp]
\centering
\caption{Detailed contrasts for the four primary evaluations. Each row identifies the changes in candidate support, objective, or search. Search limits and objective certificates are reported separately.}
\label{tab:detailed_attribution}
\small
\setlength{\tabcolsep}{4pt}
\begin{tabularx}{\textwidth}{@{}p{0.16\textwidth}p{0.19\textwidth}p{0.23\textwidth}X@{}}
\toprule
Study & Candidates and scores & Objective and constraints & Search, budget, and interpretation \\
\midrule
FLOWR: best-of-4 to re-decoding & Complete outputs versus first-output terminal-score states & Same delivery target and checks; re-decoding adds composition guidance and an edit cap & Same four generated outputs; postprocessing differs. Tests the expanded decision procedure. \\
FLOWR: local to full-domain & Same states and terminal scores & Same objective, chemistry constraints, and edit cap & Same eight one-second solve caps; only local restricts updates to four groups. Tests search scope. \\
RNA: independent/native to canonical & Same score matrices; canonical filter changes eligible edges & Pairing restrictions and native decoding rules differ & Independent selection, native flow, or matching. Tests structural decoding, with unmatched solver costs. \\
RNA: canonical to stacking & Same scores and canonical pair rules; negative edges may matter with stacking & Degree constraints shared; adds a unit stacking reward & Matching is the zero-reward objective reference. Tied optima can yield different predictions. \\
RNA: local to full-domain stack & Same candidate edges and scores & Same degree constraints and stacking objective & Up to 500 local updates versus 60-second MILP with numerical optimality status. Tests search under distinct limits. \\
DiffAb: unary to pairwise & Same amino-acid/rotamer states and unary costs & Same context filtering; adds pair terms and hard-pair fallback & Per-site minimization versus twelve coordinate sweeps. Tests pairwise modeling plus its policy. \\
DiffAb: fixed-AA control & Same identities, rotamers, and unary costs within each control & Adds pair terms; rejected pairs restore the matched unary & Archived exploratory control of rotamer coordination; its fallback differs from the primary protocol. \\
Protein: pool to recombination & Same $K$ feasible samples; complete probability rescored & Same hard constraints, reward, and KL reference & Exact support analysis with shared source budget; measures available mass and utility. \\
\bottomrule
\end{tabularx}
\end{table}

\subsection{FLOWR}

The source generator is the frozen FLOWR model of~\citet{cremer2026flowr}. The experiment uses 225 pockets from its official test file, eight random seeds per pocket, and four complete outputs per request: 1,800 requests and 7,200 source outputs. The input includes the reference-defined holo pocket and reference heavy-atom count. The evaluated task is pocket- and size-conditioned generation, using the supplied ligand size as part of the input.

Local candidates are extracted from the first output's terminal atom, charge, and bond probabilities. The other three outputs enter only the complete-output screening pool. Each heavy atom retains its two highest-scoring supported element labels; charge alternatives include neutral, raw, and top-two labels. Candidate edges include raw bonds and geometry-near pairs, with the latter defined by distance at most $1.3(r_i+r_j)+0.15$ \AA{} using the maximum covalent radius among each atom's alternatives. Bond options include no bond, single bond, the raw bond, and the top-two nonzero predictions. Atom/charge states include allowed heavy-bond valences after accounting for implicit hydrogens. Integer equations couple incident bond orders to these capacities and retain the raw total charge. These are necessary chemical constraints; connectivity and complete validity are checked after reconstruction.

Unary costs combine negative terminal log probabilities, clipped below at $10^{-12}$, with 0.1 per changed atom label, charge, or bond decision. The common edit cap is eight such changed decisions; these can span fewer than eight distinct atom sites. Coordinates remain fixed. Squared composition deviations enter the objective with a coefficient larger than the possible change in the base cost. The single target requires four nitrogen-plus-oxygen atoms; the dual target additionally requires one halogen. Each arm uses two common model-only solves followed by up to six further solves, each with a one-second limit and exclusions of previously returned assignments. Both local and full-domain arms use SciPy/HiGHS MILP formulations of the same objective; local solves additionally restrict distance from the current center to four categorical groups, while full-domain solves omit this neighborhood restriction.

Evaluation uses RDKit and PoseBusters 0.6.5 with the \texttt{mol\_fast} configuration. Best-of-four screens the four complete outputs without editing. The source package includes its aggregate outcomes and indexes the evaluator, but does not supply its ranking and tie-breaking implementation. For re-decoding, selection among proposals passing the quality checks minimizes squared composition error, then terminal regret, then observation order. If none passes, the method returns the raw first output with its recorded quality and target status. Success requires both the target composition and the specified quality checks. Overall quality pass rate is separately reported on all 1,800 requests, including unsuccessful deliveries. The comparison fixes source-generation accounting. Candidate domains and additional decoding costs follow each method. This application uses terminal-score candidate expansion; the complete-sample Cartesian support analysis in Appendix~\ref{app:d} is a distinct controlled protein experiment.

The full-domain-minus-best-of-four success differences are 49.444 percentage points (95\% pocket-cluster interval [45.167, 53.611]) for the single target and 27.556 [24.056, 31.056] for the dual target. Against local quadratic search, the differences are 0.111 [\ensuremath{-}0.167, 0.389] and 0.000 [\ensuremath{-}0.278, 0.278]. Of the selected full-domain outputs, 1,359 single-target outputs and 1,260 dual-target outputs are outside their original four-sample pools. Both counts use the archived output comparison, including target failures. The source package does not supply the graph-equivalence implementation for this comparison. On the common request denominator, the successful-and-outside count is at least $n_{\rm success}+n_{\rm outside}-1800$. This yields the bounds in Section~\ref{sec:4.1}, without requiring a claim about the absence of every feasible target hit in the pool. Pocket identity is the resampling unit, with seeds and outputs nested within each pocket.

If the delivery contract permits either arm's output, a combined policy can prioritize a verified target hit and otherwise choose a quality-passing output. Each binary success or quality event would then be the union of the corresponding events for the two arms. The reported marginals imply lower bounds of 1,309 single-target successes, 615 dual-target successes, and 1,797 quality passes for each target, out of 1,800. These are bounds for a proposed selector. Exact rates require replaying the paired output records.

The recorded four-output generation batches average 7.2146 seconds per request (median 7.0177; 95th percentile 11.6472) in the shared execution environment. Mean recorded logical solver times for local and full-domain decoding are 0.3234 and 0.3420 seconds for the single target, and 0.3502 and 0.3779 seconds for the dual target. These solver totals include the common proposal prefix. Candidate validation is cached across methods, while domain construction and final selection are timed jointly with other work; the archive therefore identifies measured components rather than isolated end-to-end method latencies. The resource transition retains 128 results without a thread-limit marker and 1,672 with a one-thread marker. The comparison in Table~\ref{tab:1} uses the shared generation budget.

Across all 1,800 first-output domains, the logged one-hot candidate count has median 493, 95th percentile 767.05, and maximum 971; categorical group counts are 64, 97, and 129, respectively. These are candidate-domain dimensions, excluding additional property auxiliaries in the MILP.

\subsection{RNA}

The frozen predictor is KnotFold~\citep{gong2024knotfold}. UFold~\citep{fu2022ufold} supplies the six public test files. Every record uses the same five-checkpoint foreground ensemble, \texttt{prior\_0.pth} through \texttt{prior\_4.pth}, and background checkpoint \texttt{reference.pth}, from KnotFold revision \texttt{cf72c56c}. Each checkpoint's paired-class softmax probability is symmetrized by averaging the two matrix directions. Let $f_{ij}$ be the mean of the five symmetrized foreground probabilities and $b_{ij}$ the symmetrized background probability. After rounding both to ten decimal places, the common score is
\[
w_{ij}=\log(f_{ij}+\epsilon)-\log(b_{ij}+\epsilon)
-\log(1-f_{ij}+\epsilon)+\log(1-b_{ij}+\epsilon)-4.2,
\qquad \epsilon=10^{-8}.
\]
The offset 4.2 is inherited from the released native decoder. Independent prediction selects all positive-score edges with positional separation at least three. Native decoding runs the released \texttt{KnotFold\_mincostflow} executable on these foreground/background matrices without a base-type filter. Canonical matching uses NetworkX maximum-weight matching on AU, UA, CG, GC, GU, and UG edges with positional separation at least three and positive scores. Nucleotides may remain unpaired, and crossing pairs are permitted.

Local and full-domain stacking share canonical edges with separation at least three and $w_{ij}\geq-2$. Both maximize the sum of selected edge scores plus a reward of 1.0 for each selected adjacent nested pair $((i,j),(i+1,j-1))$, subject to at most one partner per nucleotide. For the linear objective with zero stacking reward, deleting a negative-score edge from any feasible matching improves the objective, and omitting zero-score edges preserves its optimal value. Thus canonical matching solves the zero-reward problem on the same declared domain, up to ties. With a unit stacking reward, an edge can participate in at most two rewarded pairs. Deleting an edge with $w_{ij}<-2$ then strictly improves the stacking objective, so the cutoff preserves the optimal value. The reward and cutoff were fixed in the earlier development protocol and carried unchanged into the full evaluation. The 21 exposed TS2 records informed mechanism evaluation and the decision to expand the study. Local search starts at canonical matching and chooses the best improving move among one-edge insertions/replacements, adjacent-stem two-edge insertions/replacements, and single-edge deletions. Insertions remove incident conflicts. It stops when improvement is at most $10^{-9}$ or after 500 iterations. Full-domain inference linearizes each pair product with a binary auxiliary variable and its three standard product inequalities; a better local solution is retained if the solver provides no better incumbent.

Every source record is retained, with no length filter. The sparse MILP uses a 60-second limit per instance and requests a zero relative MIP gap. The archive reports numerical optimal termination for the 7,215 finite problems. Solver status and stored objective values concern the finite stacking objective and the solver's numerical tolerances. Macro F1 averages per-record pairing F1. Confidence intervals for paired differences use 10,000 sequence-hash cluster bootstrap resamples within each file. Duplicate sequences are clustered together. Table~\ref{tab:s1} reports provenance and paired differences for full-domain stacking versus native flow and canonical matching. The main table retains all decoding methods.

\begin{table}[t]
\centering
\caption{RNA provenance and paired macro-F1 differences for full-domain stacking versus native flow and canonical matching. Differences and pointwise 95\% sequence-hash-cluster intervals are in percentage points. The training-name count uses the released filename list.}
\label{tab:s1}
\small
\setlength{\tabcolsep}{4pt}
\begin{tabularx}{\textwidth}{@{}lrrrrXX@{}}
\toprule
File & Records & Unique & Train names & Development & Full$-$native [95\% interval] & Full$-$canonical [95\% interval] \\
\midrule
ArchiveII & 393 & 383 & 0 & 0 & 3.313 [2.877, 3.739] & 0.482 [0.141, 0.808] \\
TS0 & 1,305 & 1,305 & 714 & 0 & 0.677 [0.279, 1.073] & \ensuremath{-}1.662 [\ensuremath{-}2.028, \ensuremath{-}1.296] \\
TS1 & 60 & 60 & 0 & 0 & \ensuremath{-}0.470 [\ensuremath{-}1.570, 0.800] & 0.992 [\ensuremath{-}0.119, 2.319] \\
TS2 & 38 & 38 & 0 & 21 & 0.816 [\ensuremath{-}1.129, 3.119] & 2.622 [0.836, 4.852] \\
TS3 & 18 & 18 & 0 & 0 & 2.196 [\ensuremath{-}1.379, 6.826] & 3.583 [0.443, 7.907] \\
bpnew & 5,401 & 5,401 & 0 & 0 & 1.137 [0.978, 1.297] & \ensuremath{-}0.983 [\ensuremath{-}1.134, \ensuremath{-}0.833] \\
\bottomrule
\end{tabularx}
\end{table}

Table~\ref{tab:rna_exposure} presents the two exposure partitions using the cached predictions for every method. TS0 membership is a record-name-stem match to the released training list; TS2 development membership is a sequence-hash match to the 21 earlier development records. The partitions are exhaustive and disjoint within each file. Means retain equal record weights, and paired intervals use the same 10,000 sequence-hash bootstrap procedure. The complementary strata identify absence from these specific checks. Sequence-homology independence remains unassessed. Three ArchiveII records share sequence hashes with name-flagged TS0 records, despite having no direct name-list match themselves; the six files contain 7,197 distinct hashes in total and are reported separately.

\begin{table}[t]
\centering
\caption{RNA performance stratified by documented exposure. All F1 values are percentages; differences and 95\% sequence-hash-cluster intervals are percentage points. ``No name match'' refers only to the released training-name list, and ``outside development'' refers to the documented 21-record set.}
\label{tab:rna_exposure}
\small
\setlength{\tabcolsep}{4pt}
\begin{tabularx}{\textwidth}{@{}Xrrrrrrl@{}}
\toprule
Stratum & $n$ & Indep. & Native & Canon. & Local & Full & Full$-$native [95\% interval] \\
\midrule
TS0: training-name match & 714 & 74.140 & 78.879 & 81.409 & 79.788 & 79.605 & 0.726 [0.154, 1.331] \\
TS0: no name match & 591 & 62.327 & 67.903 & 70.010 & 68.697 & 68.520 & 0.617 [0.104, 1.143] \\
TS2: development & 21 & 78.325 & 82.637 & 81.373 & 83.165 & 83.673 & 1.036 [$-1.536$, 4.319] \\
TS2: outside development & 17 & 84.063 & 85.937 & 83.462 & 86.481 & 86.481 & 0.544 [$-2.189$, 3.579] \\
\bottomrule
\end{tabularx}
\end{table}

On the 17 TS2 records outside development, the full-domain-minus-independent difference is 2.418 percentage points [$-0.406$, 5.658], and the full-domain-minus-local difference is exactly zero. On TS0, the full-domain-minus-canonical differences are $-1.804$ [$-2.325$, $-1.260$] for name-matched records and $-1.489$ [$-1.979$, $-0.991$] for unmatched records. Canonical matching therefore ranks higher in both TS0 strata.

Table~\ref{tab:rna_dimensions} reports the recorded finite-problem dimensions and solver-call times. Each stacking term introduces one product auxiliary in the sparse MILP; the edge count is the number of native binary decisions.

\begin{table}[htbp]
\centering
\caption{RNA problem dimensions and recorded MILP solve time: median / 95th percentile / maximum within each file. All 7,215 records are represented. Solve time excludes model inference and candidate compilation; these are recorded component times. Percentiles use linear interpolation.}
\label{tab:rna_dimensions}
\small
\begin{tabular}{@{}lrrr@{}}
\toprule
File (records) & Candidate edges & Stacking terms & Solve time (s) \\
\midrule
ArchiveII (393) & 82 / 712.2 / 1583 & 47 / 401 / 977 & 0.0084 / 0.3798 / 1.8553 \\
TS0 (1,305) & 66 / 449.6 / 1583 & 41 / 273 / 977 & 0.0060 / 0.1659 / 1.9042 \\
TS1 (60) & 42.5 / 226.1 / 1065 & 23 / 134.2 / 749 & 0.0035 / 0.0467 / 1.0484 \\
TS2 (38) & 28.5 / 97.9 / 239 & 19 / 72.3 / 138 & 0.0026 / 0.0135 / 0.0474 \\
TS3 (18) & 74.5 / 289.2 / 398 & 49.5 / 177.4 / 242 & 0.0073 / 0.0755 / 0.1326 \\
bpnew (5,401) & 49 / 246 / 1348 & 32 / 153 / 878 & 0.0043 / 0.0536 / 1.4979 \\
\bottomrule
\end{tabular}
\end{table}

\subsection{Antibody CDR geometry}

The primary antibody comparison uses DiffAb~\citep{luo2022diffab}: 50 complexes with 72 nested cases each, totaling 3,600 matched cases. The 72 cases per complex combine three generation seeds (20260721, 20260722, 20260723), six target CDRs, and four samples per CDR. For each generated backbone, the amino-acid support at a design site contains the three lowest sequence-unary costs, augmented by the sampled and posterior-mode residues. PDBFixer 1.12 templates are aligned to the generated N/CA/C frame, and downstream bond-graph rotations construct at most nine deterministic rotamers per amino acid. Ala and Gly have one state; Pro retains its template ring. Ordered chi1 seeds use $-60$, $60$, and $180$ degrees, with residue-specific ordering; chi2 seeds are $90$ degrees for D/N/F/H/W/Y and $180$ otherwise. Additional unique chi1/chi2 combinations are appended and truncated at nine. The ancillary protocol sources provide the full residue-specific ordering and feature tables.

For candidate atom sets $A,B$, the steric cost is
\[
\Phi(A,B)=\sum_{p\in A,q\in B}
\left[\frac{\max(0,\rho_p+\rho_q-d_{pq}-0.4\,\text{\AA})}{0.6\,\text{\AA}}\right]^2.
\]
The unary combines negative log-softmax model scores at temperature three, CDR and curvature terms with weights 0.20 each, a fixed-context contact term with weight 0.35, and $4\Phi$ against context. Pair costs combine adjacent transition and liability terms with weights 0.18 and 0.25, generated-contact terms with weight 0.20, $2\Phi$, and a penalty of 50 for raw side-chain overlap above 1.0 \AA{}. Generated contacts require sequence separation at least three and C$\alpha$ distance at most 8 \AA{}, with Gaussian distance weighting. Fixed-context contact features use the same distance cutoff. Candidate scoring includes all fixed heavy atoms and active-CDR backbone atoms, excluding a candidate's entire own-residue backbone. The candidate radii for C, N, O, and S are 1.70, 1.55, 1.52, and 1.80 \AA{}.

The source package's \path{anc/protocol_sources/repair_diffab_antibody_qubo_outputs.py} defines these sequence features in \texttt{cdr\_prior\_cost}, \texttt{geometry\_unary}, \texttt{context\_contact\_unary}, \texttt{transition\_penalty}, \texttt{contact\_pair\_penalty}, and \texttt{developability\_pair\_penalty}. The rotamer adapter in \path{anc/protocol_sources/repair_diffab_rotamer_qubo_v2.py} applies the adjacent liability term through \texttt{add\_adjacent\_developability\_v2}. The accompanying \path{anc/protocol_sources/manifest.json} fixes the archived file hashes.

Candidates with context overlap above 1.0 \AA{} are removed for both arms. Unary selection minimizes each site's unary; pairwise selection starts there and takes at most twelve deterministic coordinate sweeps, accepting energy decreases above $10^{-12}$. Empty site supports return the packed DiffAb anchor in both arms; a selected pairwise assignment with hard pair overlap above 1.0 \AA{} also returns that anchor. Thus the primary comparison precedes the physical acceptance gate while retaining these construction-time fallback rules. It compares the complete pair model with unary selection. The archived one-hot penalty is the larger of 100 and one plus the maximum, over candidate variables, of the absolute unary coefficient plus the sum of absolute incident pair coefficients.

The four primary endpoints use placed structures and OpenMM topology. Evaluated pairs comprise CDR--context and unordered internal-CDR heavy-atom pairs, excluding atoms separated by at most three topology bonds; CDR backbone atoms are included. With $o_{pq}=\max(0,\rho_p+\rho_q-d_{pq})$ and $N$ evaluated CDR heavy atoms, the endpoints are $\sum o_{pq}^2/N$, $1000\sum\mathbf1[o_{pq}>0.4\,\text{\AA}]/N$, maximum overlap, and minimum distance. Their units and means appear below. These endpoint overlaps are unshifted; the 0.4 \AA{} threshold applies only to clash counting. The ancillary metric implementation specifies radii for every supported element, with a 1.80 \AA{} fallback, and evaluates these endpoints without a spatial truncation.

Paired differences are averaged over cases within each complex, then equally across complexes. Percentile intervals use 10,000 complex bootstrap resamples. Two-sided sign-flip tests assume symmetry of the independent complex-level differences under the null. They enumerate up to twenty nonzero differences and use 200,000 random sign assignments and the plus-one correction for larger sets. The four primary endpoints form one Holm family. The clash count per thousand atoms uses the stated heavy-atom definitions. The complex is the independent statistical unit.

\begin{table}[t]
\centering
\caption{Full CDR geometry comparison. Differences and intervals retain the precision of the paired source analysis, so subtracting rounded displayed means can differ in the last digit.}
\label{tab:s2}
\small
\setlength{\tabcolsep}{4pt}
\begin{tabularx}{\textwidth}{@{}Xrrl@{}}
\toprule
Endpoint & Unary & Pairwise & Pairwise$-$unary [95\% interval] \\
\midrule
Normalized overlap, \AA{}\textsuperscript{2}/atom & 0.327344 & 0.261613 & \ensuremath{-}0.065732 [\ensuremath{-}0.078456, \ensuremath{-}0.054008] \\
Clashes per 1,000 atoms & 320.732926 & 282.859400 & \ensuremath{-}37.873525 [\ensuremath{-}44.410927, \ensuremath{-}31.643099] \\
Maximum overlap, \AA{} & 1.399863 & 1.236854 & \ensuremath{-}0.163009 [\ensuremath{-}0.195692, \ensuremath{-}0.132723] \\
Minimum distance, \AA{} & 1.796311 & 1.932882 & 0.136571 [0.109991, 0.166171] \\
\bottomrule
\end{tabularx}
\end{table}

All four raw sign-flip p-values reach the simulation resolution of $1/(200{,}000+1)$. The resulting Holm-adjusted values are approximately $2.0\times10^{-5}$. The cohort was separated by time and exact PDB identifiers. Table~\ref{tab:cdr_exposure} further stratifies the paired results by sequence relatedness to the reconstructable SAbDab potential-training proxy. Identity is the fraction of matching shared Chothia-numbered Fv positions among comparisons with at least 80\% coverage of both sequences. Four complexes have 100\% identity over the shared positions, of which two also have complete coverage in both directions; nine more have identity from 90\% to below 100\%. These proxy strata describe relatedness, since the checkpoint-specific training list is unavailable.

\begin{table}[t]
\centering
\caption{Primary DiffAb geometry by disjoint Fv-proxy strata. Differences are pairwise minus unary, with equal-complex means and 95\% complex-bootstrap intervals. Every complex retains 72 cases. Identity uses coverage-qualified shared positions; these strata do not identify checkpoint training membership.}
\label{tab:cdr_exposure}
\small
\setlength{\tabcolsep}{4pt}
\begin{tabularx}{\textwidth}{@{}Xrrll@{}}
\toprule
Proxy identity & Complexes & Cases & $\Delta$ overlap, \AA{}\textsuperscript{2}/atom & $\Delta$ clashes/1,000 atoms \\
\midrule
100\% shared positions & 4 & 288 & $-0.12564$ [$-0.16231$, $-0.09454$] & $-69.172$ [$-87.405$, $-55.251$] \\
90\% to below 100\% & 9 & 648 & $-0.06815$ [$-0.08980$, $-0.04760$] & $-39.620$ [$-51.490$, $-28.398$] \\
Below 90\% & 37 & 2,664 & $-0.05867$ [$-0.07285$, $-0.04590$] & $-34.065$ [$-41.599$, $-27.224$] \\
\bottomrule
\end{tabularx}
\end{table}

The below-90\% intervals retain the archived sensitivity analysis. Intervals for the two complementary strata use 10,000 complex bootstrap draws; the four-complex stratum has limited independent replication. All subgroup analyses are descriptive and separate from the primary four-endpoint Holm family.

\subsubsection{Residue size and fixed-identity controls}

The variable-identity primary contrast contains 1,010 substitutions across 32,496 paired residue positions, affecting 762 of the 3,600 cases. Of these substitutions, 725 reduce side-chain heavy-atom count, 32 preserve it, and 253 increase it. The most frequent changes are Y$\to$S (167), Y$\to$G (79), D$\to$G (47), and G$\to$S (45). Side-chain counts exclude backbone atoms and hydrogens. The net count change is $-2,372$, and the mean change is $-0.65889$ atoms per CDR [$-0.78389$, $-0.54083$]. The primary contrast therefore combines residue-size changes with rotamer selection; the controls below hold identities fixed.

\begin{table}[t]
\centering
\caption{Composition and composite costs in the primary DiffAb comparison. Means average cases within complex, then weight the 50 complexes equally. Costs are archived composite energies, which include model and heuristic terms.}
\label{tab:cdr_composition}
\small
\begin{tabularx}{\textwidth}{@{}Xrrl@{}}
\toprule
Quantity per CDR & Unary & Pairwise & Difference [95\% interval] \\
\midrule
Side-chain heavy atoms & 29.54778 & 28.88889 & $-0.65889$ [$-0.78389$, $-0.54083$] \\
Placed CDR heavy atoms & 65.65444 & 64.99556 & $-0.65889$ [$-0.78333$, $-0.54167$] \\
Base composite cost & 18.31982 & 18.37356 & $+0.05374$ [0.04399, 0.06442] \\
Full candidate cost & 97.97434 & 69.01861 & $-28.95574$ [$-33.73315$, $-24.43085$] \\
\bottomrule
\end{tabularx}
\end{table}

The archive retains composite energies but no per-case logits or separated model NLL for this contrast. Table~\ref{tab:cdr_composition} therefore reports the available costs, leaving model-likelihood changes unestimated. Raw squared overlap changes by $-4.81545$ \AA{}\textsuperscript{2} per CDR, comprising an internal-CDR change of $-4.88221$ and a context change of $+0.06676$. The overall improvement includes a small increase in the context component. Among the 2,838 pairs with unchanged sequences, normalized overlap changes by $-0.02040$ [$-0.02810$, $-0.01344$]; this is a subset defined by realized outputs.

The archived fixed-sequence intervention instead holds identity constant before optimization. Both arms share the fixed sequence, backbone, context, rotamer candidates, and coefficient builder. Unary selection independently minimizes candidate unaries; pairwise selection uses the same coordinate solver with pair costs. A hard-pair rejection restores that arm's exact matched unary assignment. Sequence-only model terms are constant across the rotamer choices, and atom counts remain identical between arms.

\begin{table}[t]
\centering
\caption{Archived exploratory fixed-identity DiffAb controls. Paired geometry uses equal-complex means and the original 95\% complex-bootstrap intervals. All 50 complexes contribute. Evaluable pairs are shown against the full 3,600-case denominator; these controls are outside the primary Holm family.}
\label{tab:cdr_fixed}
\small
\setlength{\tabcolsep}{4pt}
\begin{tabularx}{\textwidth}{@{}Xrll@{}}
\toprule
Fixed identities & Evaluable/planned & $\Delta$ overlap, \AA{}\textsuperscript{2}/atom & $\Delta$ clashes/1,000 atoms \\
\midrule
Emitted DiffAb sequence & 3,600/3,600 & $-0.066829$ [$-0.075667$, $-0.058381$] & $-34.370$ [$-38.273$, $-30.455$] \\
Unary-selected sequence & 2,795/3,600 & $-0.049793$ [$-0.061750$, $-0.038884$] & $-25.482$ [$-31.126$, $-20.334$] \\
\bottomrule
\end{tabularx}
\end{table}

For emitted identities, context-feasible rotamers are retained where available; otherwise the fixed-identity raw states remain, with feasibility recorded separately. All 3,600 pairs are evaluable, including 594 hard-pair rejections restored to unary. The unary-selected-sequence control is undefined for 805 cases with an empty feasible position; they remain in its planned denominator and are excluded from paired geometry, leaving 2,795 pairs, including 185 hard-pair fallbacks. These controls demonstrate rotamer-level geometric improvements under their fixed-identity policies. Their support and fallback rules differ from the variable-identity primary protocol, so the effect sizes do not decompose its sequence and rotamer contributions. FlowPacker provides additional fixed-identity evidence in a different generator and packing cohort.

The FlowPacker comparison fixes residue identity, backbone, and non-target atoms. Each case uses four independent packing outputs, giving up to four side-chain states per residue after merging matching atom sets with coordinate RMSD at most 0.001 \AA{}. Best-of-four minimizes the same complete geometry objective. Recombination uses unary $4\Phi$ and pair $2\Phi$ plus the hard-overlap penalty of 50, with twelve local sweeps followed by a five-second SciPy MILP solve; a better local incumbent is retained. The 648 cases combine four dyMEAN samples and six CDRs for each complex.

Acceptance requires successful placement and finite physical metrics for the anchor and proposal, a changed placed structure, a decrease of at least 1.0 \AA{}\textsuperscript{2} in total squared overlap, and no increase in context squared overlap or maximum overlap. The eligible CDR--antigen contact count within 4.5 \AA{} may decrease by at most one pair, and the fraction of CDR heavy atoms with an antigen contact by at most 0.01. Rejection restores the exact best-of-four structure and metrics. The overlap margin was selected on the earlier cohort and frozen before this cohort. FlowPacker intervals use 200,000 complex bootstrap resamples and belong to a separate diagnostic family.

A stronger packing comparison uses FlowPacker~\citep{lee2025flowpacker}: 27 complexes and 648 cases, with best-of-four anchors. The protection policy accepts 35 recombined candidates and exactly falls back in 613 cases. The mean change in clashes per thousand evaluated atoms is \ensuremath{-}1.811923 [\ensuremath{-}2.8963, \ensuremath{-}0.836471]. The policy provides selective geometric refinement of the packer at a 5.40\% acceptance rate. On 145 certified optimization subinstances, local search reaches the certified optimum in 131. Geometry and optimization are the evaluated endpoints.

\FloatBarrier
\subsection{Physical CIM execution of antibody objectives}
\label{app:antibody_cim}

The hardware study transfers antibody-derived QUBOs to a physical coherent Ising machine through the Kaiwu interface. The main set comprises 50 matrices from 50 complexes, with 177--521 hardware spins. Seven enlarged matrices contain 955--2,381 original QUBO variables and 1,021--2,446 hardware spins after precision adaptation. These seven configurations derive from two source complexes, 7QUH and 9JD1. Both solvers receive the same archived symmetric signed-14-bit integer coupling matrix, with coefficient magnitudes at most 8,191. The submitted objective is to maximize $H(s)=s^\top Js$, or equivalently to minimize $-H(s)$. We evaluate the hardware objective and the original decoded molecular objective separately.

The classical comparator is Kaiwu's \texttt{FastSimulatedAnnealingOptimizer}. Each comparison uses ten retained states per solver and matrix. For the main set, FSA-10 uses the first ten rows of the archived 64-state return order; CIM uses ten selected hardware outputs. The comparison evaluates these retained candidate sets, with different physical search budgets.

We independently recompute every retained state's integer $H$ from its matrix and spin vector, then select the best value for each solver and matrix. The CIM service's energy field differs from exact $-H$ in 283/500 main-set rows and all 70 enlarged-set rows; it is excluded from ranking. The resulting raw CIM win/tie/loss counts are 48/2/0 and 7/0/0 (Table~\ref{tab:cim_antibody}). Mean normalized gaps $(H_{\rm CIM}-H_{\rm FSA})/|H_{\rm FSA}|$ are $1.7448\times10^{-5}$ and $1.2812\times10^{-7}$, respectively. These ratios describe the submitted integer objectives, including their penalty and precision-adaptation terms.

For the decoded comparison, both sets use strict restoration of split variables: inconsistent copies are rejected before restoring the original spin vector. In the enlarged set, all 70 retained CIM states and 52 of 70 FSA states pass this consistency check. Auxiliary-spin normalization and a common one-hot repair then precede up to twelve deterministic coordinate sweeps on the original molecular objective. The best hard-feasible polished result is selected from each solver's retained states. The recorded absolute tie tolerances are $10^{-9}$ for the main set and $10^{-12}$ for the enlarged set. CIM win/tie/loss counts become 0/49/1 and 1/6/0. The main-set majority-restoration sensitivity gives the same counts. Common postprocessing therefore removes most raw-objective differences in the evaluated molecular outputs.

The earlier 64-state main-set run uses a separate hardware configuration: CIM has 4 raw wins and 46 losses against FSA-64, followed by 0 wins, 48 ties, and 2 losses after majority restoration and common polishing. Its results are recorded separately from the ten-state comparison. The enlarged matrices are configuration repeats within two complexes; their seven outcomes describe a hardware workload rather than seven independent biological systems.

The main-set timing metadata contain a user-supplied device duration of 0.036464 seconds. The archived audit attributes a mean FSA-10 solver duration of 0.289000 seconds to client timing manifests. The primary CIM device receipt is absent from the retained archive, and its stated boundary excludes queueing, transfer, service overhead, and client postprocessing. The enlarged-set archive for the ten-state comparison contains no device timing record. Accordingly, Table~\ref{tab:cim_antibody} reports solution quality and hardware scale; the available records do not determine an end-to-end speed ratio. The accompanying evidence files retain per-matrix objectives, decoding summaries, input hashes, and the raw-objective recomputation record.

\FloatBarrier
\section{Additional applications}
\label{app:c}

\subsection{SemlaFlow}

SemlaFlow~\citep{irwin2025semlaflow} provides a second frozen molecular generator. The experiment balances 2,048 requests across hydrogen-inclusive node counts 24, 32, 40, and 48, with an anchor and three complete resamples per request, totaling 8,192 outputs. One evaluation exception remains in the planned denominator. The experiment evaluates a balanced size-stratified distribution over these node counts. Composition and fixed-coordinate editing rules match the declared molecular protocol; resampling uses the 512 source-generation batches as clusters.

\begin{table}[t]
\centering
\caption{SemlaFlow target-success counts out of all 2,048 requests. The final column reports verified quality passes for the full-domain method.}
\label{tab:s3}
\small
\setlength{\tabcolsep}{4pt}
\begin{tabularx}{\textwidth}{@{}Xrrrrr@{}}
\toprule
Target & Original & Best-of-4 & Local & Full & Full quality passes \\
\midrule
Single & 381 & 1,195 & 873 & 946 & 1,904 \\
Dual & 62 & 266 & 174 & 196 & 1,903 \\
\bottomrule
\end{tabularx}
\end{table}

Full-domain re-decoding improves on the original output and local search in this setting, but whole-sample screening is stronger. Under the strict terminal-regret protection variant, full-domain and local success counts both become 415 for the single target and 65 for the dual target. This contrast with FLOWR links the utility of recombination to the available states, couplings, and protection rules.

\subsection{Layout}

The frozen source is LayoutDiffusion~\citep{zhang2023layoutdiffusion}, using RICO and PubLayNet weights. The input contract keeps 8--12 elements; RICO also requires at least two Text elements. This selects 759/3,729 RICO pages and 4,905/10,998 PubLayNet pages. Two seeds and eight complete outputs per request give 1,518 and 9,810 requests, respectively. Reference coordinates are not supplied to the generator. C1 enforces the specified non-overlap relations; C2 additionally locks the largest-area anchor and allows at most four edited boxes. In RICO the non-overlap rule applies only to Text--Text pairs, not all semantic element pairs. Three PubLayNet identity-changing exceptions remain failures in the planned denominator.

The full-domain solver has a two-second limit and is compared with eight-restart local search and complete-pool selection. Open-support translation/projection or grid repair can move beyond the finite model candidate set, so it has different support. It is a local rule implementation, not the learned Layout-Corrector of~\citet{iwai2024layoutcorrector}. An earlier learned-corrector interface trial crossed coordinate quantizers and is not included in the primary ranking.

\begin{table}[t]
\centering
\caption{Layout deliveries satisfying the specified geometric contract. Denominators are 1,518 for RICO and 9,810 for PubLayNet.}
\label{tab:s4}
\small
\setlength{\tabcolsep}{4pt}
\begin{tabularx}{\textwidth}{@{}Xrrrrr@{}}
\toprule
Dataset/contract & Original & Best-of-8 & Local & Full & Open-support rule \\
\midrule
RICO/C1 & 1,358 & 1,388 & 1,406 & 1,406 & 1,517 \\
RICO/C2 & 1,358 & 1,368 & 1,405 & 1,405 & 1,516 \\
PubLayNet/C1 & 8,469 & 9,163 & 9,618 & 9,622 & 9,616 \\
PubLayNet/C2 & 8,469 & 8,491 & 9,538 & 9,551 & 9,595 \\
\bottomrule
\end{tabularx}
\end{table}

Recombination improves over complete-pool selection in these contracts, while the open-support rule often achieves higher delivery. The endpoint is satisfaction of the geometric contract; aesthetic and semantic quality are separate outcomes. Resampling clusters seeds and boxes by page hash.

\subsection{Other structured prediction tasks}

The additional interfaces expose where native specialized inference already solves the shared objective. In NMR HSQC, 434 of 479 source records are supported and contain 6,699 peak groups; 45 records have missing or ambiguous solvent information and are reported as unsupported. Direct QUBO/MIP changes independent micro accuracy from 65.726\% to 83.072\% and removes 1,985 duplicate assignments. Retaining a Hungarian anchor yields 84.356\%, identical to Hungarian matching. Peak groups are nested within molecular records.

Cell tracking covers ten two-dimensional CTC datasets, twenty sequences, and 8,017 frames. On reference detections, pooled association F1 changes from 99.74889\% to 99.77290\%, and 27 child-count violations are removed. Shared QUBO and native ILP give the same per-sequence outputs. The experiment evaluates association accuracy conditioned on reference detections; end-to-end tracking and official TRA use different evaluation inputs and endpoints. Six datasets, comprising twelve sequences, overlap the predictor's training datasets.

The three CREMI volumes contain 585,937,500 voxels, but each re-decoding neighborhood contains at most eight fragments. Local re-decoding produces the same outputs as GAEC on all three volumes. The voxel count describes image size; the fragment neighborhoods determine optimization size. For 177 ModelAngelo entries, micro residue identity changes from 54.190\% in the raw no-sequence output to 84.076\% with native HMM processing and 84.091\% with either independent register matching or shared QUBO. Coordinates remain fixed, and the additional improvement is already obtained by the local register procedure. Historical checkpoint training independence has not been certified for CREMI or ModelAngelo.

Multi-subunit assembly lacks the required effective predicted-pose support, and spatial-transcript attribution lacks confirmed independent assignment truth. These interfaces remain at the feasibility stage. The complete application source ledger records their unresolved prerequisites alongside the evaluated tasks.

\section{Probability-aware re-decoding and sampling diagnostics}
\label{app:d}

\subsection{Exact target and finite-domain enumeration}

The sequence model is ProteinMPNN~\citep{dauparas2022proteinmpnn}, checkpoint \texttt{v\_48\_020}, with coordinate noise disabled, temperature one, fixed decoding order, and an alphabet of 20 standard amino acids after removing and renormalizing X. Fixed context positions are decoded in increasing sequence order, followed by the design positions in the order listed below. Each sequence is scored under its own generated prefix; sampling probabilities and teacher-forced scores are independently compared.

For design order $\sigma=(\sigma_1,\ldots,\sigma_4)$, the conditional probability is
\[
p_0(s\mid c,\sigma)=\prod_{t=1}^{4}
\frac{p_\theta(s_{\sigma_t}\mid c,s_{\sigma_{<t}})}
{\sum_{a\in\mathcal A_{20}}p_\theta(a\mid c,s_{\sigma_{<t}})},
\]
where $\mathcal A_{20}$ is the standard-amino-acid alphabet; fixed-context likelihood factors are excluded. Across the six enumerations, the largest absolute full-domain mass residual is $8.8212\times10^{-8}$, below the recorded $5\times10^{-5}$ tolerance. Constraint-conditioned masses have maximum residual $4.4409\times10^{-16}$ (tolerance $10^{-10}$). Native-pool sampling log probabilities and enumerated scores differ by at most $4.7684\times10^{-6}$ (tolerance $2\times10^{-4}$).

The six native contexts are 1L2Y, 1LE0, 1UBQ, 1CRN, 1VII, and 2GB1. Their one-based design positions are (1,2,3,4), (1,2,3,4), (2,7,11,16), (7,8,11,12), (1,4,5,7), and (39,40,41,54). Each domain has $20^4=160,000$ states. K and R contribute +1 charge, D and E contribute \ensuremath{-}1, and all others, including histidine, contribute zero in this declared counting model. Hydrophobic residues are AVILMFWY. The hard constraints are design-region charge zero and exactly two hydrophobic residues, giving 27,648 feasible states per backbone. Aromatic residues are FWY and polar residues are STNQ; the two losses are $\ell_A=(A-1)^2/2$ and $\ell_P=(P-1)^2/2$.

Each of four independent native streams per backbone contains 4,096 draws. The first $K=4,16,64$ feasible draws form the complete pool; duplicate states are not replaced by extra draws. Recombined support is the product of the amino-acid sets observed at each site, intersected with the same hard constraints. All KL values use the complete feasible-domain reference $p_C$, including when the candidate support is smaller. Per-stream minimum KL is $-\log p_C(\mathcal S)$; averaging this quantity is not equivalent to taking the negative logarithm of average retained mass.

For reward $r$, the normalized distribution proportional to $p_C(s)\mathbf1_{\mathcal S}(s)e^{\lambda r(s)}$ gives the usual finite-domain KL-constrained variational solution when its KL equals an interior budget. This follows from nonnegativity of its KL divergence from any competing distribution. Reward increases with derivative $\operatorname{Var}(r)$, and KL with derivative $\lambda\operatorname{Var}(r)$. At maximum reward, the reference conditioned on the maximizing states gives the smallest KL on the plateau. The exact analysis uses budgets 0, 0.1, 0.25, 0.5, 1, 2, 4, and 8 nat and records infeasibility explicitly, rather than assigning zero utility. Applying these standard relations yields an exact measure of the opportunities available in each candidate support.

\begin{table}[t]
\centering
\caption{Exact dual-property target on the full feasible domain, equally weighted across six backbones. These are target-distribution quantities, not empirical MCMC estimates.}
\label{tab:s5}
\small
\setlength{\tabcolsep}{4pt}
\begin{tabularx}{\textwidth}{@{}Xrrrr@{}}
\toprule
Guidance $\lambda$ & Dual success (\%) & KL to $p_C$ (nat) & Joint TV to $p_C$ & Entropy (nat) \\
\midrule
0 & 7.3509 & 0 & 0 & 7.570392 \\
1 & 13.1567 & 0.044079 & 0.120264 & 7.695946 \\
2 & 21.8102 & 0.190252 & 0.241685 & 7.788548 \\
4 & 46.0854 & 0.777334 & 0.449763 & 7.784909 \\
\bottomrule
\end{tabularx}
\end{table}

At dual guidance four, mean site and site-pair TV are 0.196950 and 0.311249. Entropy need not decrease monotonically under property guidance because reweighting may increase the mass of previously rare states. Single guidance four yields 68.3970\% single success and 17.7503\% dual success. The target definition determines these changes. Earlier experiments on twelve- or sixteen-site domains provide independent rejection-sampling references for selected observables; the four-site enumeration evaluates the complete distribution within its finite domain.

\begin{table}[htbp]
\centering
\caption{Exact support results at $K=64$ by backbone, averaging its four source streams. Reward gain compares recombination with whole-pool support at the common KL budget of 4 nat for the dual-target reward. Streams are nested replicates.}
\label{tab:protein_backbone_support}
\small
\begin{tabular}{@{}lrrrr@{}}
\toprule
Backbone & Pool mass (\%) & Recombination mass (\%) & Reward gain & Strict-gain streams \\
\midrule
1CRN & 4.6796 & 77.7950 & 0.620710 & 4/4 \\
1L2Y & 3.9794 & 80.5786 & 0.487415 & 4/4 \\
1LE0 & 11.2977 & 74.5258 & 0.039529 & 3/4 \\
1UBQ & 7.0123 & 75.2626 & 0.534213 & 4/4 \\
1VII & 10.6920 & 72.4432 & 0.515815 & 4/4 \\
2GB1 & 50.6339 & 84.2110 & 0.138048 & 4/4 \\
\bottomrule
\end{tabular}
\end{table}

At $K=64$, minimum KL across the 24 streams ranges from 0.626912 to 3.391673 nat for whole-pool support and from 0.144648 to 0.434693 nat for recombination. Thus all recombination supports and no whole pools are feasible at 0.5 nat; both supports are feasible in every stream at 4 nat. Table~\ref{tab:protein_backbone_support} gives the corresponding context-level mass and reward results.

\subsection{Conditional preservation and property tilting}

Distributional use requires genuine joint probabilities. The analysis in this appendix uses ProteinMPNN~\citep{dauparas2022proteinmpnn} with fixed context and decoding order, temperature one, and prefixwise normalization over the 20 standard amino acids. Every recombination is rescored under its own prefixes. Nonnegative composition losses $\ell$ define

\begin{equation}
\label{eq:5}
\begin{aligned}
\pi_{\boldsymbol\beta}(s)
&=\frac{p_0(s)\mathbf1_{\mathcal C}(s)
\exp[-\boldsymbol\beta^\top\ell(s)]}{Z_{\boldsymbol\beta}},\\
&\hspace{4em}\boldsymbol\beta\geq0.
\end{aligned}
\end{equation}

Zero tilt in Eq.~\eqref{eq:5} preserves $p_C$. Positive weights favor the specified composition properties. The four-site hard constraints fix charge to zero and hydrophobic count to two. Aromatic and polar counts, $A$ and $P$, give losses $\ell_A=(A-1)^2/2$ and $\ell_P=(P-1)^2/2$; single-property and joint-property targets tilt one or both losses. Frontier rewards are $-\ell_A$ and $-(\ell_A+\ell_P)$, respectively, distinct from the indicators of target attainment. All counts concern the designed positions, with residue memberships fixed before evaluation.

Optimizer outputs can seed explicit proposals with full feasible support, followed by property-guided Metropolis--Hastings correction~\citep{emami2023ppde} using true target and proposal probabilities~\citep{hastings1970}. Production proposals are frozen after adaptation, and rejected proposals retain the current chain state. Production samples follow this transition rule without optimization acceptance filters. Proposal design governs finite-budget approximation within this fixed target family. The composition objectives specify sequence counts; protein stability and function require separate evaluation.

\subsection{Explicit proposals and probability correction}

A low-energy optimizer output seeds an explicit proposal whose probability is computed independently of the solver output law. Let $b_i(a)>0$ be smoothed frequencies estimated from an independent native calibration stream. For center $c^{(m)}$, define site factors $p_{mi}(a)=(1-\eta_m)b_i(a)+\eta_m\mathbf1[a=c_i^{(m)}]$. The proposal is

\[
q(s)=\sum_m\alpha_m\frac{\mathbf1_{\mathcal C}(s)\prod_i p_{mi}(s_i)}{Z_m},
\qquad \sum_m\alpha_m=1.
\]

Each component is conditioned and normalized before mixing. For additive charge and hydrophobic counts, suffix dynamic programming computes $B_{m,i}(v)=\sum_a p_{mi}(a)B_{m,i+1}(v-d(a))$, with $B_{m,n+1}(v)=\mathbf1[v=0]$. This provides both the normalizer and exact sequential conditional sampling for each component. A positive defensive component gives support throughout the feasible domain.

Each run pays for 128 native calibration draws and sixteen probability probes. The calibration factors are $b_i(a)=(n_i(a)+0.1)/(128+2)$, where $n_i(a)$ counts amino acid $a$ at site $i$. The mixed-center implementation uses the 171-query quadratic construction, combines its candidates with explicitly recorded conditional-sampling supplements, and selects eight centers using true target energy and sequence distance. Across 36 mixed-center runs, 92 of 288 selected center slots come from actual QUBO solutions and 196 from conditional supplements. The center enhancement coefficients are 0.25 and 0.65, with defensive mixture weight at least 0.15. The resulting proposal combines QUBO centers and conditional-sampling centers. Center selection starts with the lowest true target energy. Subsequent selections maximize the equally weighted sum of one minus normalized energy rank and minimum normalized Hamming distance to previously selected centers. Initial nondefensive mixture weights are uniform; the defensive component receives weight 0.15.

Two rounds of 512 warm-up samples adjust the site factors and mixture weights through importance weighting. Warm-up weights are proportional to the unnormalized target density divided by the density used to draw each state, then normalized over the batch. Component responsibilities under the current mixture weight each amino-acid marginal. Every nondefensive site factor becomes 0.9 times this weighted marginal plus 0.1 times its calibration factor; components with weighted mass below $10^{-10}$ retain their factors. The defensive factors stay fixed. After renormalizing the constrained components, forty responsibility updates fit mixture weights, flooring the unnormalized updates at $10^{-12}$ before normalization and enforcing defensive weight at least 0.15. This is a heuristic factor fit followed by mixture-weight fitting. Centers, factors, and weights are frozen before production. Exact enumeration tables are never read to initialize chains, select centers, or fit proposals. For feasible states, independence-MH accepts a proposal $s'$ with probability

\[
\min\left\{1,\frac{p_0(s')e^{-\boldsymbol\beta^\top\ell(s')}q(s)}{p_0(s)e^{-\boldsymbol\beta^\top\ell(s)}q(s')}\right\}.
\]

The two accepted probability flows agree, establishing detailed balance for the declared target under the stated probability and support conditions. This is the standard~\citet{hastings1970} correction. Protein-prior and property-guided MCMC also has direct precedents such as PPDE~\citep{emami2023ppde}. Product and mixed-center chains start from independent draws of their fitted proposal. Native-MH chains start from the first feasible calibration draws, cycling through them if fewer are available than chains; further native draws are taken only if none is feasible. Calibration and warm-up states are excluded from production. All rejection-induced repeated states are retained to preserve the sampling output law.

\subsection{Complete exploratory proposal comparison}

The completed diagnostic includes native autoregressive MH, conditional-product MH, and mixed-center MH: six backbones, two seeds, three target weights $(0,0)$, $(2,0)$, $(2,2)$, and three methods, totaling 108 runs with four chains each. Native proposals outside the constraints are rejected. The conditional-product implementation uses a defensive and a fitted component and the same warm-up schedule. The analysis is retained as a sampler implementation diagnostic, separate from the main comparison of complete-output screening and local candidate re-decoding.

Time begins with each run's calibration and includes probability probes, fitting, solver calls where applicable, warm-up, and production. Checkpoints are at 5, 15, and 30 seconds. Only completed batches before a checkpoint count as output. Work finishing after the final deadline is charged to cost but does not supply deadline-valid samples. Model and backbone preparation, online work, and writing costs are separately logged. Native ancestral generation and teacher-forced scoring use different computation paths and are not equated by counting each as one model call.

All 324 checkpoints produced states. The first 25\% of each chain is removed for checkpoint analysis; rejected moves remain repeated states. Absolute errors are relative to the exact target expectation. Means are equally weighted across backbones after averaging nested runs. Joint TV compares a finite empirical histogram with the exact target and contains both sampling noise and chain dependence. Unvisited mass is the exact target probability of states absent from the retained trajectory.

\begin{table}[t]
\centering
\caption{Complete proposal diagnostic. Each row contains 36 runs. Success errors are measured in percentage points relative to the exact target expectations.}
\label{tab:s6}
\small
\setlength{\tabcolsep}{4pt}
\begin{tabularx}{\textwidth}{@{}lXrrrr@{}}
\toprule
Budget & Proposal & Single MAE & Dual MAE & Joint TV & Unvisited mass \\
\midrule
5 s & Native MH & 3.793 & 2.525 & 0.7483 & 0.7012 \\
5 s & Conditional product MH & 1.258 & 0.820 & 0.4261 & 0.3485 \\
5 s & Mixed-center MH & 2.153 & 1.924 & 0.5142 & 0.4516 \\
15 s & Native MH & 2.284 & 1.975 & 0.6499 & 0.5810 \\
15 s & Conditional product MH & 0.900 & 0.705 & 0.3043 & 0.2164 \\
15 s & Mixed-center MH & 2.080 & 1.413 & 0.3558 & 0.2709 \\
30 s & Native MH & 1.326 & 1.579 & 0.5796 & 0.4972 \\
30 s & Conditional product MH & 0.824 & 0.581 & 0.2466 & 0.1581 \\
30 s & Mixed-center MH & 1.517 & 1.220 & 0.2882 & 0.2008 \\
\bottomrule
\end{tabularx}
\end{table}

Conditional-product proposals have lower errors than mixed-center proposals in this diagnostic. The mixed-center proposal has thirty-second site and site-pair TV of 0.03965 and 0.09466, compared with 0.02720 and 0.07046 for the conditional-product proposal. Full cold-start median costs, including preparation and writing, are 35.743, 35.810, and 35.797 seconds for native, product, and mixed-center implementations. Preparation takes 3.621--6.017 seconds, and the largest final in-flight overrun is 0.264 seconds. These timings include the measured overheads in the shared-GPU environment.

\subsection{Invariance, endpoint bounds, and finite trajectories}

ESS, Monte Carlo standard error, and rank-normalized $\hat R$ are computed for applicable nonconstant success indicators, with the diagnostic distinctions of~\citet{vehtari2021rhat}. These diagnostics characterize the monitored observables; exact finite-domain calculations additionally quantify coverage of the target states. On the finite domain, $\delta=\min_{s\in\mathcal C}q(s)/\pi(s)>0$ yields the independence-MH minorization bound $(1-\delta)^t$ for a deterministic number of steps $t$. Native proposals retain their infeasible-domain rejection probability in this calculation; they are not silently renormalized to $p_C$.

At deterministic step counts numerically equal to the observed counts, the largest native fixed-step bound is $1.6180\times10^{-9}$. This is a fixed-step calculation. A random wall-clock stopping time requires a separate analysis, even when its realized step count is recorded. The native trajectories have empirical joint TV of 0.5796 at 30 seconds. This statistic compares a finite correlated trajectory histogram with the target, whereas the theoretical bound concerns the law of one endpoint at a specified deterministic step.

At the 30-second checkpoint, the mixed-center proposal has median ESS 1520.8 for single-target success and 1856.4 for dual-target success, over 36 saved run indicators per endpoint. Corresponding median MCSE values are 1.042 and 0.610 percentage points, and maximum rank-normalized $\hat R$ values are 1.0490 and 1.0811. All indicators at this checkpoint are nonconstant. Across all saved checkpoints and proposals, one of 648 indicators is constant and is excluded from these diagnostics. These monitored statistics describe the recorded trajectories and do not certify coverage of unvisited modes.

\section{Retrospective selection using public GB1 measurements}
\label{app:e}

The source measurements are the GB1 four-site landscape of~\citet{wu2016gb1}, processed in the official ProteinGym v1.3 data~\citep{proteingymdata} as \texttt{SPG1\_STRSG\_Wu\_2016}. Of 149,360 measured variants, 26,675 satisfy the design-region hard constraints. GB1 sites 39, 40, 41, and 54 map to full annotation positions 265, 266, 267, and 280. Unmeasured variants and an absent wild-type row are not assigned zero activity.

The fixed context uses the annotated 56-residue DQ-starting fragment on the 2GB1 backbone template. Its probabilities are rescored separately from the native MT-starting context of Appendix~\ref{app:d}. The context follows the benchmark annotation. The study retrospectively evaluates policies on the recorded variants; equivalence to the complete physical assay construct remains unverified.

Eleven policies are frozen before reading outcome values: preserve plus single and dual guidance at strengths 0.25, 0.5, 1, 2, and 4. Their support is the measured feasible subset, which retains 99.8910\% of the full hard-conditional model mass. The query distribution is the uniform mixture $r$ of these policies. Each queried variant reveals one existing measurement, and policy $q_j$ is evaluated with weight $q_j/r\leq11$. For each policy, the estimator averages the products of these weights and binary outcomes over all query draws, including repeated variants. Selection maximizes this unnormalized importance-weighted average, with ties resolved by the first policy in the listed order. Sampling is with replacement; repeated queries do not count as additional distinct labels. Two hundred query streams use budgets 8--256. The complete labels are consulted after policy freezing to evaluate population utility and regret, rather than to construct the policies.

The primary label is the provided \texttt{DMS\_score\_bin}, using its median threshold 0.1224388752. Positive-label probability measures performance relative to the benchmark cutoff. Continuous \texttt{DMS\_score} is supplementary. Policy construction requires 160,000 model scores, taking 77.721 seconds in the observed shared-GPU environment; this cost is separate from label budget.

\begin{table}[t]
\centering
\caption{Policy selection from public GB1 outcomes, averaged over 200 query streams from one assay.}
\label{tab:s7}
\small
\setlength{\tabcolsep}{4pt}
\begin{tabularx}{\textwidth}{@{}Xrrr@{}}
\toprule
Draw budget & Distinct labels & Expected binary score & Regret to best policy \\
\midrule
8 & 7.585 & 0.883658 & 0.021957 \\
16 & 14.320 & 0.887004 & 0.018611 \\
32 & 25.885 & 0.888038 & 0.017578 \\
64 & 44.250 & 0.893146 & 0.012469 \\
128 & 71.825 & 0.896124 & 0.009491 \\
256 & 111.745 & 0.900631 & 0.004984 \\
\bottomrule
\end{tabularx}
\end{table}

Preserve has expected score 0.894519; the best frozen policy is dual guidance one, with 0.905615. Mean selected utility at budgets 8--64 does not exceed preserve. At 256 draws, the gain is approximately 0.611 percentage points. Strong single guidance four instead gives 0.823080 and strong dual guidance four gives 0.871368. Continuous expected scores are 0.901837 for preserve, 0.907088 for dual guidance one, and 0.825696 for single guidance four. Composition guidance is therefore distinct from assay utility. The observed policy-selection benefit depends on query budget and guidance strength within this assay.

\section{Reproducibility and evidence scope}
\label{app:g}

Source artifacts retain frozen protocols, model identities, candidate pools, objective matrices, solver states, random seeds, probabilities, and accepted/rejected transitions. Primary outcomes retain planned denominators and declared independent units. Main-task data, exact-support tables, sampler diagnostics, and public-label results are bound to their source files in the companion \texttt{source\_manifest.txt} included in the LaTeX source package. This manifest records source-file hashes; the underlying experimental assets remain separate from the LaTeX source package.

The evidence combines staged application studies into one evaluation of the shared re-decoding interface. Exact support calculations are reserved for analysis and evaluation. Archived antibody code and later task adapters implement one common mathematical abstraction through separate implementations. Independent source checks and trajectory replay verify numerical consistency. The task protocols specify biological endpoints, training provenance, and distributional diagnostics. Credentials and private caches are excluded from the manuscript and artifact index.

The ancillary directory \path{anc/protocol_sources/} supplies the archived candidate and feature-table definitions, antibody geometry evaluators, RNA decoder operations, and protein proposal and policy-estimation routines, with source hashes. These protocol sources document the implementations used here; full experiments additionally require their model weights, input datasets, and runtime dependencies.

\setlength{\parskip}{3pt}
\section{Peptide alphabet and topology extensions}
\label{app:peptide_extensions}

These QUBO capability studies apply the candidate, constraint, and reconstruction interface to noncanonical substitution, cyclization, peptide-grammar decoding, and fixed-chemistry pose assembly. External templates specify new chemical identities or topology; model outputs provide family preferences, atom-slot information, or linear conformations. The pose-assembly example uses traditional docking proposals. Design assignments, generation requests, source pools, and closure starts are reported with their own denominators and acceptance rules.

\begin{table}[htbp]
\centering
\caption{Peptide capability settings. These studies verify different chemical or geometric outputs under their own candidate domains and budgets.}
\label{tab:peptide_attribution}
\small
\setlength{\tabcolsep}{4pt}
\begin{tabularx}{\textwidth}{@{}p{0.16\textwidth}p{0.19\textwidth}p{0.23\textwidth}X@{}}
\toprule
Study & Candidates and scores & Objective and constraints & Search, budget, and interpretation \\
\midrule
Linear-model cyclization & Supplied sequence; 32 sampled linear poses per pool & Cyclic graph, fixed anchors, closure and geometry checks & QUBO results over four pools, with 32 closure starts each. Tests geometric cyclization capability. \\
Noncanonical substitutions & External chemistry/rotamers and backbone-only model scores & Exactly two substitutions; physical and family-preference terms & All 960 assignments enumerated; eight QUBO designs reported. Tests controlled alphabet extension. \\
Peptide grammar & Frozen FlowMol3 slots/scores; external G/A/S/V templates & Exact atom coverage and peptide topology; soft editing costs & Per-template mapping QUBOs; chemistry and two separate coordinate routes checked. Tests specification delivery. \\
Thioether poses & Fixed Vina pose fragments; source-score and geometric costs & One state per block; cross-block bond and clash restrictions & All 20,736 assignments enumerated; full geometry checks follow. Tests fragment reconstruction. \\
\bottomrule
\end{tabularx}
\end{table}

\subsection{Noncanonical double substitutions on a deposited PUMA backbone}
\label{app:ncaa}

The input is deposited PUMA--MCL1, PDB 2ROC model 1, with receptor chain A, peptide chain B, deposited free termini, and original protonation. Editable author-numbered positions B149, B152, B142, B144, B141, and B137 were selected by native side-chain contacts within 4.5~\AA{} of the receptor, excluding terminal residues, Gly/Pro, and incomplete local backbones. This structure differs from the capped PUMA27 assay construct at residue 144 and the C terminus. The experiment evaluates the deposited structure without using numerical assay responses.

The external chemical library uses Rosetta's NVL and HLU residue definitions and backbone-dependent rotamers~\citep{renfrew2012ncaa}. Each position retains one parent state and the four highest-probability rotamers at the nearest periodic backbone-angle grid for each noncanonical identity. All receptor coordinates, the peptide backbone, and nonedited residues remain fixed; no continuous refinement is applied. The six groups contain 54 one-hot bits. Exactly two new noncanonical substitutions give $\binom{6}{2}8^2=960$ conformer assignments and $\binom{6}{2}2^2=60$ chemical sequences. Existing single-substitution chemical checks precede this enumeration.

Frozen ProteinMPNN~\citep{dauparas2022proteinmpnn}, checkpoint \texttt{v\_48\_020}, supplies backbone-only unconditional canonical scores. The receptor backbone remains in context; probabilities are renormalized after removing X. Mapping $g(\mathrm{NVL})=\mathrm{V}$ and $g(\mathrm{HLU})=\mathrm{L}$ gives the substitution cost
\[
c_i(a)=\log p_i(a_i^0)-\log p_i(g(a)).
\]
Rotamers of the same chemical identity share this family preference. Input placeholder identities A, W, and X produce identical backbone-only scores. The mapping transfers a canonical side-chain-family preference to an external state; it does not define a noncanonical sequence probability or extend the reference distribution of Section~\ref{sec:2.3}.

Physical differences retain the original \texttt{mm\_std} weights. Each residue identity uses the minimum intraresidue score among its four existing rotamers as a finite-library strain reference, relative to the parent's intraresidue score. The complete objective adds family cost with a fixed coefficient of one Rosetta energy unit (REU) per nat. A fixed receptor histidinium torsion lacks an engine parameter, so the calculation cancels invariant fixed-residue terms instead of assigning an absolute total energy. All terms involving an edited residue are evaluated, and changes to fixed--fixed hydrogen bonds are explicitly included when a substitution changes their environment context. This correction applies to 160 double assignments, with maximum magnitude 0.10881 REU. On representative single- and double-substitution checks, comparison with all fixed--fixed context-dependent terms gives error below $10^{-7}$ REU. On fully parameterized 1L2Y single- and double-substitution checks, the same difference calculation agrees with full energy differences to $9.10\times10^{-13}$ REU. Histidinium protonation and torsion weights are unchanged. These are fixed-structure score differences with a finite-library reference, rather than binding free energies.

Single and double physical evaluations define unary costs and anchored finite differences as in Appendix~\ref{app:a}. Because the allowed domain has exactly two changes, the resulting quadratic table represents every allowed assignment exactly, including the evaluated environmental changes. This interpolation property does not imply pairwise additivity for larger edit sets. Squared one-hot and substitution-count penalties, together with forbidden candidate pairs, use the sufficient bound $2S+1$. Enumeration checks all 960 assignments; direct and quadratic soft costs differ by at most $4.55\times10^{-13}$ REU, and the complete penalized polynomial by $1.17\times10^{-10}$ REU.

Verification reconstructs the complete peptide graph and checks valence, connectivity, the two-substitution count, residue parameter graphs, all peptide bonds, coordinate-derived stereochemistry, and preservation of fixed atoms. Every covalent heavy-atom distance must lie within 1.10--1.90~\AA{}. Severe nonbonded contacts use a 1.8~\AA{} threshold, excluding peptide atom pairs at graph distance at most two. Receptor contacts use the same threshold. All 960 assignments pass the graph, stereochemistry, and covalent-distance checks; 510 fail the receptor-contact threshold, leaving 450 qualified conformers across all 60 chemical sequences. These counts describe the coverage of one candidate library, not independent design requests.

\begin{table}[t]
\centering
\caption{Eight distinct noncanonical designs selected by QUBO re-decoding on the deposited PUMA--MCL1 backbone. Each design has exactly two new substitutions and one selected conformer. Physical values use the finite-library strain reference; the QUBO soft objective adds family cost at one REU/nat. Qualification requires all declared chemical and geometric checks.}
\label{tab:ncaa_extensions}
\small
\setlength{\tabcolsep}{5pt}
\begin{tabularx}{\textwidth}{@{}rXrrrr@{}}
\toprule
Rank & Substitutions & Physical (REU) & Family cost (nat) & Objective (REU) & Qualified \\
\midrule
1 & B142:HLU, B144:NVL & -6.8190 & 4.1059 & -2.7131 & Pass \\
2 & B142:HLU, B152:NVL & -7.2086 & 4.9253 & -2.2833 & Pass \\
3 & B141:NVL, B142:HLU & -8.1620 & 7.2067 & -0.9554 & Pass \\
4 & B142:NVL, B144:NVL & -5.1826 & 6.2826 & 1.1000 & Pass \\
5 & B142:NVL, B152:NVL & -5.5722 & 7.1021 & 1.5299 & Pass \\
6 & B141:NVL, B142:NVL & -6.5256 & 9.3834 & 2.8579 & Pass \\
7 & B141:HLU, B142:HLU & 6.2654 & 2.0176 & 8.2830 & Pass \\
8 & B141:HLU, B142:NVL & 7.9018 & 4.1943 & 12.0962 & Pass \\
\bottomrule
\end{tabularx}
\end{table}

Table~\ref{tab:ncaa_extensions} lists the eight selected chemical sequences. The top-ranked design is B142:HLU/B144:NVL, with physical cost $-6.8190$ REU, family cost 4.1059 nat, and combined objective $-2.7131$ REU. Enumeration certifies the optimum over the declared qualified finite domain. The scores describe fixed-structure selection; experimental affinity is not evaluated.

Independently parsed SDF and all-atom complex PDB files for these eight designs preserve original fixed coordinates and pass graph, coordinate-derived stereochemistry, and contact checks. Broken side-chain connectivity, reflected stereochemistry with stale tags, duplicated atom mappings, and a three-dimensional broken bond with an intact graph are all rejected. Candidate preparation, enumeration and output-verification records are retained with the archived artifacts. The demonstrated capability is incorporation of the specified external chemical states under a fixed-backbone design contract.

\subsection{Cyclization of frozen linear-peptide model outputs}
\label{app:cyclization}

Frozen PepGLAD~\citep{kong2024pepglad} uses its fixed-sequence checkpoint to sample linear conformations of \texttt{LEDGWVDIETGKE} in the 9HGD/GABARAP receptor environment. The deposited holo receptor supplies coordinates; the pocket is a 44-residue neighborhood within 10~\AA{} of six specified receptor hotspots, measured using C$\beta$ or C$\alpha$ for glycine. Native source preprocessing permits receptor movement under soft restraints; compilation, closure, and refinement restore and fix the deposited receptor. Reference peptide coordinates are isolated from generation, fragment selection, and closure, and used only for final RMSD evaluation. Four new source pools, each with 32 planned samples, were locked before generation and all completed without replacing failures or changing scientific settings. Source seeds are 531201--531232, 531301--531332, 531401--531432, and 531501--531532. These pools, and their anchor conditions, are technical repeats of one sequence--receptor system.

Four generated anchor conditions are selected per pool using an anchor--receptor proxy and whole-peptide C$\alpha$ diversity. Each source pose is aligned to the chosen anchor's residue 6--8 backbone, and all anchor heavy-atom coordinates are retained from that selected source. The mobile blocks cover residues 1--3, 4--5, 9--10, and 11--13. QUBO re-decoding selects a source conformation for each mobile block under the chosen anchor, allowing blocks from different source poses to be combined. Each block has eight alternatives, giving 32 one-hot bits and $8^4=4,096$ assignments per anchor. The quadratic objective combines receptor/anchor costs, pairwise splice and nonbonded terms, and forbidden pairs. All 16 problems are enumerated; 782--1,935 assignments per problem satisfy the discrete constraints. The top eight per anchor are sent to the common closure procedure. Exact minimization establishes the proxy optimum; only the selected starts receive continuous closure and refinement.

Selected combinations receive the head-to-tail cyclic chemical graph and undergo Generalized Kinematic Closure (GenKIC)~\citep{mulligan2021genkic}, followed by constrained refinement. QUBO re-decoding has at most 32 selected starts per pool, 16 GenKIC attempts per start, and 200 refinement iterations. Qualification checks all covalent geometry, amide geometry, coordinate-derived stereochemistry, severe internal/receptor contacts, and the fixed region. The primary pose endpoint additionally requires receptor-aligned heavy-atom RMSD $\leq2$~\AA{}, with no separate ligand alignment. Closure counts include every budgeted start.

The reconstructed cyclic graph contains 104 heavy atoms, 106 bonds, and 13 stereocentres. Generic bond lengths must lie within 0.70--1.25 times the sum of covalent radii. Each amide requires a C--N distance of 1.20--1.50~\AA{}, adjacent angles of 105--135 degrees, and a plane deviation at most 25 degrees from either cis or trans. Severe internal or receptor contacts use the 1.8~\AA{} threshold, excluding intrapeptide graph distances at most two. Fixed heavy coordinates are checked at $10^{-6}$~\AA{} precision. Final Top-1 is the qualified output with lowest post-refinement \texttt{ref2015\_cart} score including position restraints; job identity breaks ties. This deployment score, not reference RMSD, selects the final output.

\begin{table}[t]
\centering
\caption{QUBO cyclization results for four new source pools of one supplied 9HGD/GABARAP sequence. Qualified counts require all declared chemical and geometric checks after closure and refinement. Denominators are closure starts. The final column additionally requires receptor-aligned heavy-atom RMSD $\leq2$~\AA{}.}
\label{tab:cyclization_extensions}
\small
\begin{tabular}{@{}lrrr@{}}
\toprule
New source pool & Closure starts & Geometry-qualified & Qualified RMSD $\leq2$~\AA{} \\
\midrule
1 & 32 & 0 & 0 \\
2 & 32 & 2 & 0 \\
3 & 32 & 1 & 0 \\
4 & 32 & 0 & 0 \\
\midrule
Total & 128 & 3 & 0 \\
\bottomrule
\end{tabular}
\end{table}

QUBO fragment selection followed by closure yields three geometrically qualified cyclic conformations (Table~\ref{tab:cyclization_extensions}). Their RMSDs are 11.086, 10.537, and 8.263~\AA{}. They share the supplied chemical sequence and are distinct conformations. Two of four source pools yield a qualified top-ranked accepted output. None reaches the primary valid, RMSD $\leq2$~\AA{} binding-pose endpoint.

Earlier QUBO development and first-confirmation pools are kept separate from this locked repetition: their qualified counts were 2/32 and 0/32 starts, respectively, with zero near-native outputs in both. The four new pools were completed as one predetermined group. Replaying the two earlier qualified QUBO examples reproduces their coordinates, scores, and acceptance decisions exactly; replay contributes no independent samples.

Across the four new pools, QUBO-selected starts allocate 128 closure/refinement jobs, with up to 2,048 GenKIC attempts and 200 refinement iterations per job. Every saved structure is independently reread with no acceptance disagreement. Candidate matrices, source membership, closure failures, and complete run-cost records are indexed in the accompanying peptide evidence manifest. The fixed-sequence experiment evaluates linear-conformation sampling followed by cyclization; it does not evaluate end-to-end design of a new sequence with the model's sequence--structure codesign checkpoint.

\subsection{Peptide-grammar decoding of general molecular outputs}
\label{app:peptide_grammar}

Frozen FlowMol3~\citep{dunn2025flowmol3}, with its default GEOM-Drugs checkpoint, supplies 32 cached unconditional trajectories. Generation receives an all-atom slot budget and random seed; peptide sequence, length, and connectivity enter the decoder. Requested sizes are 32, 40, 48, and 56 slots, each used eight times. Terminal sampled identities determine retained heavy atoms; hydrogen and fake slots are excluded. The same 32 endpoints receive two specifications: neutral linear tripeptides with free termini, and neutral head-to-tail cyclic tetrapeptides. Both use Gly/Ala/Ser/Val, with L stereochemistry at nonglycine $\alpha$ carbons. The endpoint checkpoint, code revision, seeds, scores, and slot identities are hash-bound in the ancillary records.

For each endpoint with $N$ heavy atoms, the external grammar enumerates every size-compatible sequence, deduplicating cyclic rotations. Each sequence defines a template graph with $N$ atom roles. A separate QUBO chooses the role-to-slot bijection; sequence choice is an outer enumeration over these template problems. With $z_{ia}=1$ assigning template atom $i$ to original slot $a$, the objective is
\begin{equation}
\label{eq:peptide_assignment}
E(z)=E_{\mathrm{slots}}(z)+P\sum_i\Bigl(\sum_a z_{ia}-1\Bigr)^2
                         +P\sum_a\Bigl(\sum_i z_{ia}-1\Bigr)^2.
\end{equation}
The $N^2$ bits in Eq.~\eqref{eq:peptide_assignment} enforce exact coverage of the retained source slots without additional heavy atoms. Unary costs combine endpoint element and neutral-charge log-score differences with one unit per changed element. Pair terms combine endpoint bond-category log-score differences with 0.5 units per changed category, including no-bond, over every unordered slot pair. A constant no-bond baseline plus bonded-template corrections counts each slot pair once under a bijection. Target covalent bonds additionally contribute $0.1[(d-r_0)/0.25]^2$, where distances are in \AA{} and $r_0$ is 1.33 for an interresidue amide, 1.23 for C=O, 1.43 for a remaining O bond, 1.47 for a remaining N bond, and 1.53 for C--C. Probabilities are clipped at $10^{-12}$ for local scoring. These are terminal-state preferences, separate from a complete molecular probability.

This configuration allows all required physical categories in the checkpoint vocabulary and uses soft editing costs. Earlier development imposed hard element and bond edit limits of $\lceil0.2N\rceil$ and $\lceil0.5N\rceil$; the present capability results use the expanded permissions. At most four distinct graphs per request are checked in objective order. The first initialization minimizes unary assignment cost, followed by three random permutations. Each start receives 4,000 permutation-swap annealing proposals, with temperature decreasing from 5 to 0.05, and the best result receives at most 100 improving swaps. The search preserves both assignment constraints. The exported pure QUBO uses $P=2S+1$, with $S$ the upper-triangular soft coefficient absolute sum. These runs provide feasible assignments without global-optimality certificates.

Independent graph reconstruction checks the exact peptide topology, residue count, neutral termini, and requested stereochemistry against a separately built sequence library. Two coordinate routes then undergo distinct checks. The fresh route performs 16 seeded ETKDGv3 attempts per graph and up to 500 MMFF94s minimization iterations per attempt. The source route performs one MMFF94s minimization from the model's original heavy coordinates, using soft positional restraints with 0.5~\AA{} tolerance and force constant 100. Qualification requires minimizer convergence, every covalent distance within 0.75--1.25 times the covalent-radius sum, no nonbonded distance below 0.65 times the van der Waals-radius sum beyond graph distance two, and the specified stereochemistry inferred anew from coordinates. Amide dihedrals are recorded diagnostically. The soft source restraint permits motion beyond its tolerance.

\begin{table}[t]
\centering
\caption{QUBO peptide-grammar capability on the same 32 cached FlowMol3 endpoints. Requests of incompatible size remain in the denominator. Fresh coordinates are generated by ETKDGv3/MMFF94s; the source route relaxes original model coordinates. Each route reports any qualified result among at most four checked graphs per request. Unique delivered graphs count the first qualified fresh-route result per request.}
\label{tab:peptide_grammar}
\small
\begin{tabularx}{\textwidth}{@{}Xrrrrr@{}}
\toprule
Specification & Size eligible & Fresh delivery & Source delivery & Unique delivered & QUBO bits \\
\midrule
Linear tripeptide & 19/32 & 19/32 & 2/32 & 17 & 169--484 \\
Cyclic tetrapeptide & 28/32 & 28/32 & 3/32 & 22 & 256--729 \\
\bottomrule
\end{tabularx}
\end{table}

All size-compatible requests produce a qualified fresh-route result (Table~\ref{tab:peptide_grammar}); each request's rank-one graph already qualifies in that route. The source-route successes occur at ranks two through four. None of the 155 selected graph instances passes the specified geometry checks at its unrelaxed source coordinates. Source-route qualified products have heavy-atom RMS displacements of 0.798--0.968~\AA{} for linear peptides and 0.993--1.160~\AA{} for cyclic peptides. For the first fresh-route deliveries, median element/bond edit counts are 6/11 and 6/13, respectively. Two linear and one cyclic delivery also satisfy both earlier hard edit limits. These outcomes establish grammar-directed reconstruction under the current editing permissions.

The linear and cyclic specifications require 143 and 160 template QUBOs, respectively. All 155 fresh-route SDF files are independently reread and pass the declared checks; five accepted source-route structures are also exported and reread. Independent saved-assignment QUBO energy checks differ by at most $1.41\times10^{-8}$. Five targeted implementation tests cover template construction, swap increments, and polynomial consistency. Eight CPU workers complete decoding and verification in 23.75 seconds of incremental wall time, excluding the cached generator calls. The archived sources distinguish template chemistry, QUBO-selected mapping, and the coordinate procedure used for each output.

\subsection{Fixed-chemistry thioether fragment reconstruction}
\label{app:thioether_feasibility}

The 5KEZ case applies the interface to a chemically fixed thioether peptide using traditional docking proposals. Its graph has 89 heavy atoms and 95 bonds, with deposited modifications, stereochemistry, and the thioether connection retained. Receptor chain A is fixed with Ca and Cl; waters are removed. The 40~\AA{} cubic search region is centered on the mean C$\alpha$ coordinates of receptor residues Asp197, Glu233, and Asp300, selected from an independent active-site annotation. Peptide reference coordinates are isolated from proposal generation, candidate selection, and objective construction.

The cached pool contains 32 AutoDock Vina 1.2.7 poses~\citep{trott2010vina}, seeds 510001--510032. Each starts from graph-only ETKDGv3/MMFF94s initialization and uses exhaustiveness one, 20,000 maximum evaluations, one output pose, and the same receptor. Four blocks own residues $\{1,2,6\}$, $\{3,4,5\}$, $\{7,8\}$, and $\{9,10,11\}$, respectively; the first contains the bridge region. Each heavy atom belongs to one block. Candidate fragments pass local chemistry/geometry checks, are ordered by native source score, deduplicated below 0.5~\AA{} block RMSD using graph-valid symmetries, and capped at 12 states per block. This produces 48 one-hot bits and $12^4=20,736$ assignments.

Unary costs combine source Vina affinity weighted by the fraction of atoms owned by a block, its internal bond-length proxy, and fixed-receptor interactions. The source term transfers a support score to fragments; it is not a rescoring of the assembled pose. With covalent radii $r_i^{\rm cov}$, a target bond contributes $[(d_{ij}-r_i^{\rm cov}-r_j^{\rm cov})/0.2]^2$. Each ligand--receptor heavy-atom pair contributes $10[0.75(r_i^{\rm vdw}+r_j^{\rm vdw})-d_{ij}]_+^2-0.02\exp[-((d_{ij}-3.8)/0.8)^2]$. Pairwise terms include the same bond proxy across blocks and repulsion $10[0.7(r_i^{\rm vdw}+r_j^{\rm vdw})-d_{ij}]_+^2$ for cross-block pairs beyond graph distance two. All distances are in \AA{}, and $r_{ij}^{\rm cov}=r_i^{\rm cov}+r_j^{\rm cov}$. Hard forbidden pairs exclude cross-block covalent distances outside $[0.6r_{ij}^{\rm cov},r_{ij}^{\rm cov}+0.6]$ or nonbonded distances below 0.65~\AA{}. These expressions define a geometric selection proxy.

The QUBO uses one-hot squared penalties and 453 forbidden-pair terms, each weighted by $P=2S+1$. Complete enumeration certifies the finite proxy optimum and identifies 18 assignments satisfying the encoded restrictions. All 20,736 one-hot polynomial evaluations agree within $1.87\times10^{-9}$; 258 arbitrary bit strings additionally verify the penalty expansion within $2.39\times10^{-7}$. Direct full-coordinate proxy calculations agree within $2.85\times10^{-14}$. The lowest-energy assignment fails the final protein-distance check. The acceptance policy tests assignments in energy order and delivers the first passing result, at rank two.

Full verification checks the indexed chemical graph, label and coordinate-derived stereochemistry, all covalent bonds, severe overlaps, a 1.4--2.2~\AA{} thioether distance, and the frozen PoseBusters 0.6.5 \texttt{dock\_fast} configuration~\citep{buttenschoen2024posebusters}; its energy-ratio test is disabled. Of 18 checked assignments, 13 qualify, including two that combine fragments from different source poses (Table~\ref{tab:thioether_feasibility}). Reconstruction performs no continuous refinement. All saved SDF files reproduce their acceptance status on readback. A broken-bridge counterexample with an unchanged chemical graph fails the coordinate checks.

\begin{table}[t]
\centering
\caption{Verified 5KEZ outputs in QUBO energy order. The accepted delivery retains one complete source; two later qualified outputs combine sources. All rows pass the declared chemical and geometric checks. RMSD uses the fixed receptor frame and graph-valid symmetry mappings, without separately aligning the ligand.}
\label{tab:thioether_feasibility}
\small
\begin{tabularx}{\textwidth}{@{}lrXrr@{}}
\toprule
Output & Rank & Source seeds for the four blocks & Full RMSD (\AA{}) & Bridge RMSD (\AA{}) \\
\midrule
Accepted delivery & 2 & 510019, 510019, 510019, 510019 & 10.775 & 10.768 \\
Qualified recombination & 12 & 510020, 510020, 510026, 510026 & 11.990 & 12.887 \\
Qualified recombination & 14 & 510004, 510004, 510004, 510019 & 10.951 & 12.363 \\
\bottomrule
\end{tabularx}
\end{table}

The selection record is committed before reference-coordinate evaluation. No qualified output reaches the full-peptide RMSD $\leq2$~\AA{} endpoint. This one-target study establishes QUBO fragment reconstruction with an intact modified-peptide graph and verified geometry; the finite proxy ranks and the measured pose errors describe separate outcomes. Compilation, enumeration, and verification take 34.47 seconds of incremental wall time, with the original pose-generation cost retained in the cached source records. The ancillary package includes the complete QUBO, forbidden pairs, candidate provenance, selection/evaluation records, and the two mixed-source SDF outputs.

\end{document}